\documentclass[%
 reprint,
superscriptaddress,
 amsmath,amssymb,
 aps,
onecolumn
]{revtex4-2}

\usepackage{microtype}
\usepackage{graphicx}
\usepackage{subfigure}
\usepackage{booktabs} 
\usepackage{bbm}
\usepackage{xcolor}
\usepackage{enumitem}

\usepackage{hyperref}

\usepackage{amsmath}
\usepackage{amssymb}
\usepackage{mathtools}
\usepackage{amsthm}
\usepackage{xfrac}

\usepackage[capitalize,noabbrev]{cleveref}

\theoremstyle{plain}
\newtheorem{theorem}{Theorem}[section]

\theoremstyle{definition}

\newtheorem{assumption}[theorem]{Assumption}
\theoremstyle{remark}
\newtheorem{remark}[theorem]{Remark}

\newcommand\sign{\mathrm{sign}}
\newcommand\eig{\omega}
\def\dd{\text{d}}

\newcommand{\hc}[1]{{\color{black} #1}}
\newcommand{\scaling}[1]{{\color{black} #1}}
\newcommand{\revisionL}[1]{{\color{black} #1}}
\newcommand{\revision}[1]{{\color{black} #1}}

\begin{document}
\title{Error Scaling Laws for Kernel Classification under Source and Capacity Conditions }

\author{Hugo Cui}
\affiliation{Statistical Physics Of Computation Laboratory, \'Ecole Polytechnique F\'ed\'erale de Lausanne (EPFL)
}%
\author{Bruno Loureiro}
\affiliation{Information Learning and Physics Laboratory, \'Ecole Polytechnique F\'ed\'erale de Lausanne (EPFL)
}%
\author{Florent Krzakala}%
\affiliation{Information Learning and Physics Laboratory, \'Ecole Polytechnique F\'ed\'erale de Lausanne (EPFL)
}%
\author{Lenka Zdeborov\'a}
\affiliation{Statistical Physics Of Computation Laboratory, \'Ecole Polytechnique F\'ed\'erale de Lausanne (EPFL)
}%

\begin{abstract}
In this manuscript we consider the problem of kernel classification. While worst-case bounds on the decay rate of the prediction error with the number of samples are known for some classifiers, they often fail to accurately describe the learning curves of real data sets. In this work, we consider the important class of data sets satisfying the standard source and capacity conditions, comprising a number of real data sets as we show numerically. Under the Gaussian design, we derive the decay rates for the misclassification (prediction) error as a function of the source and capacity coefficients. We do so for two standard kernel classification settings, namely margin-maximizing Support Vector Machines (SVM) and ridge classification, and contrast the two methods. We find that our rates tightly describe the learning curves for this class of data sets, and are also observed on real data. Our results can also be seen as an explicit prediction of the exponents of a scaling law for kernel classification that is accurate on some real datasets.
\end{abstract}
\maketitle

\section{Introduction and related work}
\label{sec:intro}

\scaling{A recent line of work \cite{hestness2017deep,kaplan2020scaling,rosenfeld2019constructive,henighan2020scaling}
has empirically evidenced that the test error of neural networks often obey scaling laws with the number of parameters of the model, training set size, or other model parameters. Because of their implications in terms of relating performance and model size, these findings have been the object of sustained theoretical attention. Authors of \cite{sharma2022scaling} relate the decay rate of the test loss with the number of parameters to the intrinsic dimension of the data. This idea is refined by \cite{bahri2021explaining} for the case of regression tasks, building on the observation that in a number of settings, the covariance of the learnt features exhibits a power-law spectrum, whose rate of decay controls the scaling of the error. This investigation is actually very closely related to another large body of works. In fact, the study of a power-law features spectrum (and of a target function whose components in the corresponding eigenbasis also decay as a power-law) has a long history in the kernel literature, dating back to the seminal works of \cite{caponnetto2007optimal, Caponnetto2005FastRF}. The corresponding rates governing the power-laws are respectively known as the \textit{capacity} and \textit{source} coefficients, and the scaling of the test error with the training set size can be entirely characterized in terms of these two numbers. While the study of kernel ridge regression \cite{caponnetto2007optimal,Caponnetto2005FastRF,Lin2018OptimalRF,Jun2019KernelTR,liu2020kernel,pillaud2018statistical,Berthier2020TightNC,Varre2021LastIC,Cui2021GeneralizationER} therefore offers a rich viewpoint on the question of neural scaling laws with the training set size, little is so far known for kernel \textit{classification}. Since ascertaining the test error decay under source and capacity conditions would automatically translate into neural scaling laws in \textit{classification} tasks -- similarly to \cite{bahri2021explaining} for regression -- this is a question of sizeable interest addressed in the present work.
}

 \subsection*{Related works}

\scaling{\paragraph{Neural scaling laws --- } 
A number of works \cite{hestness2017deep,kaplan2020scaling,rosenfeld2019constructive,henighan2020scaling} have provided empirical evidence of scaling laws in neural networks, with the number of parameters, training samples, compute, or other observables. These findings motivated theoretical investigations of the underlying mechanisms. Authors of \cite{sharma2022scaling} show how the scaling of the test loss with the number of parameters is related to the intrinsic dimension of the data. This dimension is further tied in with the kernel spectrum by \cite{bahri2021explaining}, a work that leverages the kernel ridge regression viewpoint to translate, in turn, the decay of the spectrum to test error rates. Authors of \cite{maloney2022solvable} similarly study a simple toy model where the power-law data is processed through a random features layer. Finally, \cite{hutter2021learning} investigate a toy model of scalar integer data in the context of classification, and ascertain the corresponding scaling law. Relating in classification settings the rate of decay of the kernel spectrum to the test error, like \cite{bahri2021explaining} for regression, is still an open question.
}\\
 
\paragraph{Source and capacity conditions ---} The source and capacity conditions are standard regularity assumptions in the theoretical study of kernel methods, as they allow to subsume a large class of learning setups, c.f. \cite{MarteauFerey2019BeyondLF,pillaud2018statistical,Caponnetto2005FastRF, caponnetto2007optimal, Cui2021GeneralizationER,Berthier2020TightNC}.\\

\paragraph{Kernel ridge regression ---}The error rates for kernel ridge regression have been extensively and rigorously characterized in terms of the source/capacity coefficients in the seminal work of \cite{Caponnetto2005FastRF,caponnetto2007optimal}, with a sizeable body of work being subsequently devoted thereto \cite{steinwart2009optimal,Lin2018OptimalRF,Jun2019KernelTR,liu2020kernel,pillaud2018statistical,Berthier2020TightNC,Varre2021LastIC}. In particular, in \cite{Cui2021GeneralizationER} it was shown that rates derived under worst-case assumptions \cite{Lin2018OptimalRF, Jun2019KernelTR, Caponnetto2005FastRF, caponnetto2007optimal, Bartlett2020BenignOI} are identical to the typical rates computed under the standard Gaussian design \cite{dobriban2018high,dicker2016ridge,Hsu2012} assumption. Crucially, it was observed that many real data-sets satisfy the source/capacity conditions, and display learning rates in very good agreement to the theoretical values \cite{Cui2021GeneralizationER}.\\

\paragraph{Worst-case analyses for SVM ---} The worst-case bounds for Support Vector Machines (SVM) classification \scaling{-- see e.g. \cite{Steinwart2008SupportVM,scholkopf2002learning} for general introductions thereto--} are known from the seminal works of \cite{Steinwart2008SupportVM,Steinwart2007FastRF,Audibert2007FastLR}. However, it is not known how tightly the corresponding rates hold for a given realistic data distributions, \revisionL{not even for synthetic Gaussian data. We show that, contrary to the case of ridge regression, for classification the worst case bounds are not tight for Gaussian data. }This effectively hinders the ability to predict and understand the error rates for relevant classes of data-sets, and in particular the class of data described by source/capacity conditions, which as mentioned above includes many real data-sets \cite{Cui2021GeneralizationER}. The key goal of this work is to fill this gap \revisionL{by leveraging the recent work on learning curves for the Gaussian covariate model \cite{Loureiro2021CapturingTL} specified to data satisfying the capacity and source conditions.}

\subsection*{Main contribution } In this work, we investigate the decay rate of the misclassification (generalization) error for \hc{}{noiseless} kernel classification, under the Gaussian design and source/capacity regularity assumptions with capacity coefficient $\alpha$ and source coefficient~$r$. Building on the analytic framework of \cite{Loureiro2021CapturingTL}, we consider the two most widely used classifiers: margin-maximizing Support Vector Machines (SVMs) and ridge classifiers. 
We derive in Section \ref{sec:MM} the error rate (describing the decay of the prediction error with the number of samples) for margin-maximizing SVM : 
\begin{align*}
\epsilon_g^{\rm SVM} \sim n^{- \frac{\alpha\mathrm{min}(r,\frac{1}{2})}{1+\alpha\mathrm{min}(r,\frac{1}{2})}}.
\end{align*}
\hc{As a consequence, we conclude that the worst-case rates \cite{Steinwart2007FastRF,Steinwart2008SupportVM,Audibert2007FastLR} are indeed loose and fail to describe this class of data}. \revisionL{This fact alone is not at all surprising. However, it becomes remarkable in the light of the fact that for ridge regression the worst case bounds and the typical case rates do agree \cite{Cui2021GeneralizationER}.}\\

We contrast the SVM rate with the rate for optimally regularized ridge classification, which we establish in Section \ref{sec:ell2} to be 
\begin{align*}
\epsilon_g^{\rm ridge} \sim n^{-\frac{ \alpha\mathrm{min}(r,1)}{1+2\alpha\mathrm{min}(r,1)}}.
\end{align*}
We argue in the light of these findings that the SVM always displays faster rates than the ridge classifier for the classification task considered. \\

Finally, we observe that some real data-sets fall in the same universality class as the considered setting, in the sense that, as illustrated in Section \ref{sec:real}, their error rates are in very good agreement with the ones above. This work is thus a key step for theoretically predicting the error rates of kernel classification for a broad range of real data-sets.

\section{Setting}
\label{sec:setting}

\subsection{Kernel classification}
Consider a data-set $\mathcal{D}=\{(x^\mu,y^\mu)\}_{\mu=1}^n$ with $n$ independent samples from a probability measure $\nu$ on $\mathcal{X}\times \{-1,+1\}$, with $\mathcal{X}\subset \mathbb{R}^d$. 
We will assume that 
the labels can be expressed as 
\begin{equation}
y^\mu=\sign{}(f^\star(x^\mu)) \label{eq:gt_function}
\end{equation}
for some \revision{non-stochastic} target function $f^\star:\mathcal{X}\rightarrow \mathbb{R}$. \revision{Note that the \textit{noiseless} setting considered here is out of the validity domain of many worst case analyses, whose bounds become void without noise \cite{Audibert2007FastLR}, whereas a number of real learning settings are well described by a noiseless setup, see section \ref{sec:real}.} Learning to classify $\mathcal{D}$ in the direct space $\mathcal{X}$ for a \textit{linear} $f^\star$ has been the object of extensive studies. 
In the present work, we focus on the case where $f^\star$ more generically belongs to the space of square-integrable functions $L^2(\mathcal{X})$. To classify $\mathcal{D}$, a natural method is then to perform \textit{kernel classification} in a $p$- dimensional Reproducing Kernel Hilbert Space (RKHS)  $\mathcal{H}$ associated to a kernel $K$, by minimizing the regularized empirical risk: 
\begin{equation}
\label{eq:generic_risk}
    \hat{\mathcal{R}}_n(f)=\frac{1}{n}\sum\limits_{\mu=1}^n\ell(f(x^\mu),y^\mu)+\lambda ||f||^2_{\mathcal{H}}.
\end{equation}
The function $\ell(\cdot)$ is a loss function and $\lambda$ is the strength of the $\ell_2$ regularization term. 
In this paper we shall more specifically consider the losses $\ell(z,y)=\mathrm{max}(0,1-yz)$ (hinge classification) and $\ell(z, y)=(y-z)^2$ (ridge classification), and the case of an infinite dimensional RKHS ($p=\infty$). The risk \eqref{eq:generic_risk} admits a dual rewriting in terms of a standard parametric risk. To see this, diagonalize $K$ in an orthogonal basis of \textit{kernel features} $\{\psi_k(\cdot)\}_{k=1}^p$ of $L^2(\mathcal{X})$, with corresponding eigenvalues $\{\eig_k\}_{k=1}^p$:
\begin{equation}
    \int_{\mathcal{X}}\nu(\dd{x}')K(x,x')\psi_k(x')=\eig_k\psi_k(x).
    \label{eq:def_psi}
\end{equation}
It is convenient to normalize the eigenfunctions to 
\begin{equation}
    \int_{\mathcal{X}}\nu(\dd{x})\psi_k(x)^2=\eig_k,
\end{equation}
so that the kernel $K$ can be rewritten in simple scalar product form $K(x,x')= \psi(x)^\top\psi(x')$, where we named $\psi(x)$ the $p$-dimensional vector with components $\{\psi_k(x)\}_{k=1}^p$. 
 
Furthermore, note that the covariance $\Sigma$ of the data in feature space with this choice of feature map is simply diagonal
\begin{equation}
\label{eq:def_Sigma}
     \Sigma=\mathbb{E}_{x\sim\nu}(\psi(x)\psi(x)^{\top})=\mathrm{diag}(\eig_1,\cdots,\eig_p).
\end{equation}
 Any function $f\in \mathcal{H}$ can then be expressed as $f(\cdot)=w^{\top}\psi(\cdot)$ for a vector $w$ with square summable components. Using this parametrization, the risk \eqref{eq:generic_risk} can be rewritten as
\begin{equation}
\label{eq:parametric_risk}
    \hat{\mathcal{R}}_n(w)=\frac{1}{n}\sum\limits_{\mu=1}^n\ell(w^{\top}\psi(x^\mu),y^\mu)+\lambda w^{\top}w.
\end{equation}
Throughout this manuscript we will refer to the components of the target function in the features basis as the \textit{teacher} $\theta^\star$, so that $$f^\star(\cdot)=\theta^{\star \top}\psi(\cdot).$$ 
\hc{Note that any $f^\star\in L^2(\mathcal{X})$ can be formally written in this form with a certain $\theta^\star$ (allowing for non square-summable components if $f^\star\in L^2(\mathcal{X})\setminus \mathcal{H}$).}
Similarly, the minimizer $\hat{w}$ of the parametric risk \eqref{eq:parametric_risk} is related to the argmin $\hat{f}$ of \eqref{eq:generic_risk} by $\hat{f}(\cdot)=\hat{w}^\top\psi(\cdot)$, and will be referred to as the \textit{estimator} in the following. We make two further assumptions : first, we work under the \textit{Gaussian design}, and assume the features $\psi(x)$ to follow a Gaussian distribution with covariance $\Sigma$, i.e. $\psi(x)\sim\mathcal{N}(0,\Sigma)$. Note that this assumption might appear constraining \scaling{, as the distribution of the data in feature space strongly depends on its distribution in the original space, and the feature map associated to the kernel. In fact, for a large class of data distributions and standard kernels, the Gaussian design assumption does not hold. However,} rates derived under Gaussian design can hold more broadly. For instance, the rates established by \cite{Cui2021GeneralizationER} under Gaussian design were later proven by \cite{jin2021learning} under weaker conditions on the features. We will moreover discuss in Section \ref{sec:real} several settings in which our theoretical rates are in good agreement with rates observed for real data. \\
 
Second, following \cite{Cui2021GeneralizationER}, we assume that the \textit{regularization strength $\lambda$ decays as a power-law} of the number of samples $n$ with an exponent $\ell $: $\lambda=n^{-\ell}$. Note that this form of regularization is natural, since the need for regularizing is lesser for larger training sets. Furthermore, this allows to investigate the classical question of the \textit{asymptotically optimal regularization} \cite{Caponnetto2005FastRF,caponnetto2007optimal,Cui2021GeneralizationER}, i.e. the decay $\ell$ of the regularization yielding fastest decrease of the prediction error.

\subsection{Source and capacity conditions}
Under the above assumptions of Gaussian design with features covariance $\Sigma$ and existence of a teacher $\theta^\star$ that generates the labels using eq.~(\ref{eq:gt_function}) we can now study the error rates.
In statistical learning theory one often uses the \textit{source and capacity conditions}, which assume the existence of two parameters $\alpha>1, r\ge 0$ (hereafter referred to as the \textit{capacity coefficient} and the \textit{source coefficient} respectively) so that
\begin{align}
\label{eq:source_capa_conditions}
    \mathrm{tr}\Sigma^{\frac{1}{\alpha}}<\infty, && \theta^{\star\top}\Sigma^{1-2r}\theta^\star<\infty.
\end{align}
As in \cite{dobriban2018high,spigler2019asymptotic, bordelon2020, Berthier2020TightNC,Cui2021GeneralizationER}, we will consider the particular case where both the spectrum of $\Sigma$ and the teacher components $\theta^\star_k$ have exactly a power-law form satisfying the limiting source/capacity conditions \eqref{eq:source_capa_conditions}:
\begin{align}
    \eig_k=k^{-\alpha}\, ,&&\theta^\star_k=k^{-\frac{1+\alpha (2r-1)}{2}}\, .
 \label{eq:model_def}
\end{align}
The power-law forms \eqref{eq:model_def} have been empirically found in \cite{Cui2021GeneralizationER} in the context of kernel regression to be a reasonable approximation for a number of real data-sets including MNIST \cite{lecun1998gradient} and Fashion MNIST \cite{xiao2017fashion} and a number of standard kernels such as polynomial kernels and radial basis functions. Similar observations were also made in the present work and are discussed in \ref{App:real} and section~\ref{sec:real}.

The capacity parameter $\alpha$ and source parameter $r$ capture the complexity of the data-set \scaling{in feature space -- i.e. after the data is transformed through the kernel feature map into $\{\psi(x^\mu),y^\mu\}_{\mu=1}^n$}. A large $\alpha$, for example, signals that the spectrum of the data covariance $\Sigma$ displays a fast decay, implying that the data effectively lies along a small number of directions, and has a low effective dimension. Conversely, a small capacity $\alpha$ means that the data is effectively large dimensional, and therefore a priori harder to learn. Similarly, a large $r$ signals a good alignment of the teacher $\theta^\star$ with the main directions of the data, and a priori an easier learning task. In terms of the target function $f^\star$, larger $r$ correspond to smoother $f^\star$. Note that $r>\sfrac{1}{2}$ implies that $f^\star\in\mathcal{H}$, while $r\le\sfrac{1}{2}$ implies $f^\star\in L^2(\mathcal{X})\setminus\mathcal{H}$. Finally, note that while \cite{MarteauFerey2019BeyondLF} suggested an alternative definition for the source and capacity coefficients in the case of non-square loss functions, their redefinition is not directly applicable for the hinge loss.

\subsection{Misclassification error}

The performance of learning the data-set $\mathcal{D}$ using kernel classification \eqref{eq:parametric_risk} is quantified by the misclassification (generalization) error
\begin{equation}
\label{eq:error}
    \epsilon_g=\frac{1}{2}-\frac{1}{2}\mathbb{E}_{\mathcal{D}}\mathbb{E}_{x,y\sim\nu}\left(y~\sign{}(\hat{w}^{\top}\psi(x))\right),
\end{equation}
where $\hat{w}$ is the minimizer of the risk \eqref{eq:parametric_risk}. The error \eqref{eq:error} corresponds to the probability for the predicted label $\sign{}(\hat{w}^{\top}\psi(x))$ of a test sample $x$ to be incorrect.
The rate at which the error \eqref{eq:error} decays with the number of samples $n$ in $\mathcal{D}$ depends on the complexity of the data-set, as captured by the source and capacity coefficients $\alpha, r$ eq.~\eqref{eq:model_def}. To compute this rate, we build upon the work of \cite{Loureiro2021CapturingTL} who, following a long body of work in the statistical physics literature \cite{mezard1987spin,Dietrich1999StatisticalMO,engel2001statistical,mezard2009information,bordelon2020,advani2020high}, provided and proved a mathematically rigorous closed form asymptotic characterization of the misclassification error as 
\begin{align}
    \epsilon_g=\frac{1}{\pi}\mathrm{arccos}\left(\sqrt{\eta}\right), &&\eta=\frac{m^2}{\rho q},
\end{align}
where $\rho$ is the squared $L^2(\mathcal{X})$ norm of the target function $f^\star$, i.e.
$
    \rho=\int_{\mathcal{X}}\nu (\dd{x}) f^{\star}(x)^2=\theta^{\star\top}\Sigma\theta^\star,
$ and $m,q$ are the solution of a set of self-consistent equations, which are later detailed and analyzed in Section \ref{sec:MM} for margin-maximizing SVMs and section \ref{sec:ell2} for ridge classifiers. The order parameters $m,q$ are known as the \textit{magnetization} and the \textit{self-overlap} in statistical physics and respectively correspond to the target/estimator and estimator/estimator $L^2(\mathcal{X})$ correlations:
\begin{align}
\label{eq:m_q_physical_meaning}
    m=\mathbb{E}_{\mathcal{D}}\int_{\mathcal{X}}\nu (\dd{x}) f^{\star}(x)\hat{f}(x)=\mathbb{E}_{\mathcal{D}}\left(\hat{w}^\top\Sigma\theta^\star\right),&&q=\mathbb{E}_{\mathcal{D}}\int_{\mathcal{X}}\nu (\dd{x}) \hat{f}(x)^2=\mathbb{E}_{\mathcal{D}}\left(\hat{w}^\top\Sigma\hat{w}\right).
\end{align}
It follows from these interpretations that $\eta$ has to be thought of as the \textit{cosine-similarity} between the teacher $\theta^\star$ and the estimator $\hat{w}$, with perfect alignment ($\eta=1$) resulting in minimal error $\epsilon_g=0$ from \eqref{eq:error}.\\ 

Note that while this characterization has formally been proven in \cite{Loureiro2021CapturingTL} in the asymptotic proportional $n,p\rightarrow \infty,\sfrac{n}{p}=\mathcal{O}(1)$ limit, we are presently using it in the $n\ll p = \infty$ limit, thereby effectively working at $\sfrac{n}{p}=0^+$. The non-asymptotic rate guarantees of \cite{Loureiro2021CapturingTL} are nevertheless encouraging in this respect, although a finer control of the limit would be warranted to put the present analysis on fully rigorous grounds. Further, \cite{Cui2021GeneralizationER} also build on \cite{Loureiro2021CapturingTL} in the $\sfrac{n}{p}=0^+$ limit, and display solid numerics-backed results, later rigorously proven by \cite{jin2021learning}. We thus conjecture that this limit can be taken as well safely in our case. {\color{black} Finally, we mention that a recent line of works \cite{li2021statistical,ariosto2022statistical,seroussi2023separation,cui2023optimal} has explored the connections between kernel regression and Bayesian learning for networks in the $\sfrac{n}{p}=\mathcal{O}(1)$ limit, where $p$ is in this case the width of the network. While the high-dimensional limit is indeed related to the one originally discussed in \cite{Loureiro2021CapturingTL}, which we relax here to $\sfrac{n}{p}=0^+$, the main object of \cite{li2021statistical,ariosto2022statistical,seroussi2023separation} was not to study kernel regression per se, but to show how observables in Bayesian regression could be expressed in terms of well-chosen kernels. In the present work, we focus on analyzing kernel classification in the $\sfrac{n}{p}=0^+$ regime.}

\section{Max-margin classification}
\label{sec:MM}

\begin{figure}
\centering
    \includegraphics[scale=0.44]{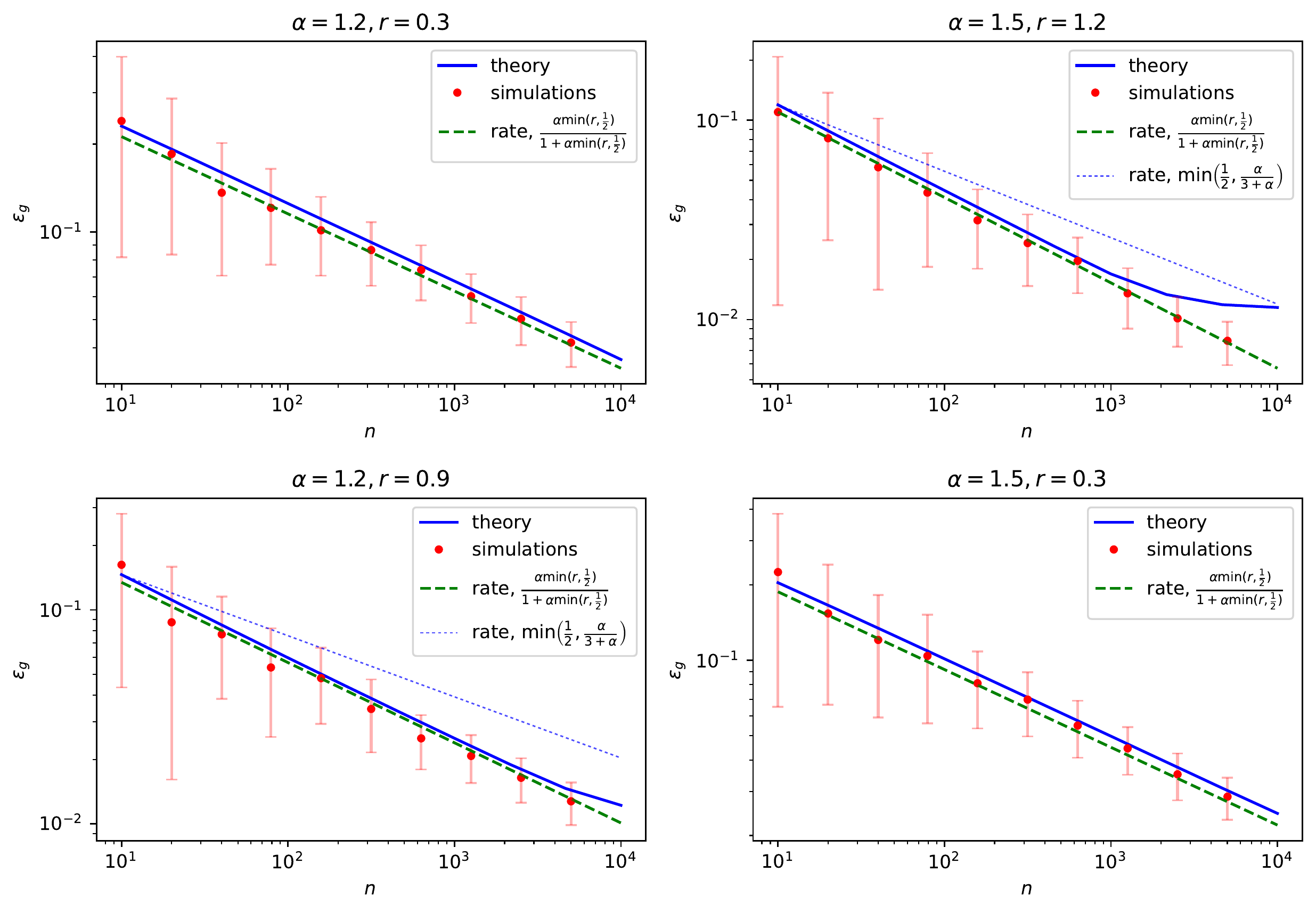}
    \caption{Misclassification error $\epsilon_g$ for max-margin classification on synthetic Gaussian features, as specified in \eqref{eq:model_def}, for different source/capacity coefficients $\alpha,r$. In blue, the solution of the closed set of eqs.~\eqref{eq:g3m_mm} used in the characterization \eqref{eq:error} for the misclassification error, using the \texttt{g3m} package \cite{Loureiro2021CapturingTL}. The dimension $p$ was cut-off at $10^4$. Red dots corresponds to simulations using the \texttt{scikit-learn SVC} \cite{pedregosa2011scikit} package run for vanishing regularization $\lambda=10^{-4}$ and averaged over $40$ instances, for $p=10^4$. The green dashed line indicates the power-law rate \eqref{eq:scaling_mm} derived in this work. \hc{The light blue dotted line indicates the classical worst-case $\min\left(\sfrac{1}{2},\sfrac{\alpha}{(3+\alpha)}\right)$ rate for SVM classification (Theorem $2.3$ in \cite{Steinwart2008SupportVM}) in the cases where the theorem readily applies ($r>\sfrac{1}{2}$)} (see also \ref{App:loose}). The code used for the simulations is available  \href{https://github.com/SPOC-group/Kernel_classification}{here.}}
    \label{fig:mm_error}
\end{figure}
\subsection{Self-consistent equations}
In this section we study regression using Support Vector Machines. The risk \eqref{eq:parametric_risk} then reads for the hinge loss
\begin{equation}
\label{eq:MM_risk}
    \hat{\mathcal{R}}_n(w)=\frac{1}{n}\sum\limits_{\mu=1}^n\mathrm{max}\left(0,1-y^\mu w^{\top}\psi(x^\mu)\right)+\lambda w^{\top}w.
\end{equation}
In the following, we shall focus more specifically on the max-margin limit with $\lambda=0^+$. We show in \ref{App:reg} that zero regularization is indeed asymptotically optimal for the data following eq.~\eqref{eq:model_def} when the target function is characterized by a source $r\le\sfrac{1}{2}$, i.e. $f^\star\in L^2(\mathcal{X})\setminus \mathcal{H}$. We heuristically expect margin maximization to be \textit{a fortiori} optimal also  for easier and smoother teachers $f^\star\in\mathcal{H}$.  For the risk \eqref{eq:MM_risk} at $\lambda=0^+$, the self-consistent equations defining $m,q$ in \eqref{eq:m_q_physical_meaning} read (see \ref{App:mm})
  
\begin{align}\label{eq:g3m_mm}
\begin{cases}
\rho=&\sum\limits_{k=1}^\infty\theta_k^{\star 2}\eig_k,\\
m=&\hat{r}_1\frac{n}{z}\sum\limits_{k=1}^\infty\frac{\eig_k^2\theta_k^{\star 2}}{1+\frac{n}{z}\eig_k},\\
q=&\hat{r}_1^2\frac{n^2}{z^2}\sum\limits_{k=1}^\infty\frac{\theta_k^{\star 2}\eig_k^3}{(1+\frac{n}{z}\eig_k)^2}\\
&+\hat{r}_2\frac{n}{z^2}\sum\limits_{k=1}^\infty\frac{\eig_k^2}{(1+\frac{n}{z}\eig_k)^2} 
\end{cases},
&&
\begin{cases}
\hat{r}_1=&\frac{1}{2\pi\sqrt{\rho}} 
\frac{\left(
\sqrt{2\pi}(1+\mathrm{erf}(\frac{1}{\sqrt{2q(1-\eta)}}))+2
e^{-\frac{1}{2q(1-\eta)}}\sqrt{q(1-\eta)}
\right)}{\int\limits_{-\infty}^{\frac{1}{\sqrt{q}}}dx\frac{e^{-\frac{1}{2}x^2}}{\sqrt{2\pi}} \left[1+\mathrm{erf}(\sqrt{\frac{\eta}{2(1-\eta)}}x)
 \right]},\\
\hat{r}_2=&\frac{\int\limits_{-\infty}^{\frac{1}{\sqrt{q}}}dx\frac{e^{-\frac{1}{2}x^2}}{\sqrt{2\pi}} \left[1+\mathrm{erf}(\sqrt{\frac{\eta}{2(1-\eta)}}x)
 \right](1-\sqrt{q}x)^2}{\left(\int\limits_{-\infty}^{\frac{1}{\sqrt{q}}}dx\frac{e^{-\frac{1}{2}x^2}}{\sqrt{2\pi}} \left[1+\mathrm{erf}(\sqrt{\frac{\eta}{2(1-\eta)}}x)
 \right]\right)^2},\\
 z=&\frac{\frac{z}{n}\sum\limits_{k=1}^\infty \frac{\eig_k}{\frac{z}{n}+\eig_k}}{\int\limits_{-\infty}^{\frac{1}{\sqrt{q}}}dx\frac{e^{-\frac{1}{2}x^2}}{\sqrt{2\pi}} \left[1+\mathrm{erf}(\sqrt{\frac{\eta}{2(1-\eta)}}x)
 \right]}. 
\end{cases}
\end{align}

Here $\hat{r}_1$ should be thought of as the ratio between the norms of the estimator $\hat{w}$ and the teacher $\theta^\star$, while $z$ can be loosely interpreted as an effective regularization. A detailed derivation of these equations can be found in \ref{App:mm}.

\subsection{Decay rates for max-margin}
From the investigation of the eqs.~\eqref{eq:g3m_mm}, as detailed in \ref{App:mm}, the following scalings are found to hold between the order parameters:
\begin{equation}
\label{eq:mm_op_scaling}
    m\sim \sqrt{q}\sim\hat{r}_1\sim n\left(\frac{z}{n}\right)^{\frac{1}{\alpha}}\sim n^{\frac{\alpha\mathrm{min}(r,\frac{1}{2})}{1+\alpha\mathrm{min}(r,\frac{1}{2})}}.
\end{equation}
 Note that the mutual scaling between $m,~q$ also follows intuitively from the interpretation of these order parameters -- \scaling{ as the overlap of $\hat{w}$ with the ground truth and itself respectively --  see the discussion around eqs. }\eqref{eq:m_q_physical_meaning} and \eqref{eq:g3m_mm}. Since the width of the margin is generically expected to shrink with the number of samples (as more training data are likely to be sampled close to the separating hyperplane), the increase of the norm of $\hat{w}$ (as captured by $q,\hat{r}_1$) with $n$ is also intuitive. Finally, an analysis of the subleading corrections to $m$ and $q$, detailed in \ref{App:mm}, leads to 
\begin{equation}
\label{eq:scaling_mm}
    \epsilon_g\sim n^{-\frac{\alpha\mathrm{min}(r,\frac{1}{2})}{1+\alpha\mathrm{min}(r,\frac{1}{2})}}.
\end{equation}

The error rate \eqref{eq:scaling_mm} stands in very good agreement with numerical simulations on artificial Gaussian features generated using the model specification \eqref{eq:model_def}, see Fig.\,\ref{fig:mm_error}. Two observations can further be made on the decay rate \eqref{eq:scaling_mm}. First, the rate is as expected an increasing function of $\alpha$ (low-dimensionality of the features) and $r$ (smoothness of the target $f^\star$). Second, for a source $r>\sfrac{1}{2}$ (corresponding to a target $f^\star\in \mathcal{H}$), the rate saturates, suggesting that all functions in $\mathcal{H}$ are all equally easy to classify, while for rougher target $f^\star\in L^2(\mathcal{X})\setminus\mathcal{H}$ the specific roughness of the target function, as captured by its source coefficient $r$, matters and conditions the rate of decay of the error.\\

\revision{Finally, we briefly discuss for completeness in \ref{App:Steinwart} the more general case where the label distribution \eqref{eq:gt_function} includes data noise, and show that the rates display a crossover from the noiseless value \eqref{eq:scaling_mm} to a noisy value, much like what was reported for kernel ridge regression \cite{Cui2021GeneralizationER}.}

\hc{
\revision{
\subsection{Comparison to classical rates}
To the best of the authors' knowledge, there currently exists little work addressing the error rates for datasets satisfying source and capacity conditions \eqref{eq:source_capa_conditions}. The closest result is the worst-case bound of \cite{Steinwart2008SupportVM} for SVM classification, which can be adapted to the present setting provided $f^\star\in\mathcal{H}$ ($r>\sfrac{1}{2}$). The derivation is detailed in \ref{App:loose} and results in an upper bound of $\min\left(\sfrac{1}{2},\sfrac{\alpha}{(3+\alpha)}\right)$ for the error rate for max-margin classification, which is always \textit{slower} than \eqref{eq:scaling_mm}. This rate \cite{Steinwart2008SupportVM} is plotted for comparison in Fig.\,\ref{fig:mm_error} against numerical simulations and is visibly off, failing to capture the learning curves. It is to be expected that the worst case rates will be loose when compared to rate that assume a specific data distribution. What makes our result interesting is the comparison with the more commonly studied ridge regression where, as discussed already in the introduction, the worst case rates actually match those derived for Gaussian data, see \cite{Cui2021GeneralizationER}. \\

Importantly, the rates from \cite{Steinwart2008SupportVM} only hold for capacity $r>\sfrac{1}{2}$, while real datasets are typically characterized by sources $r<\sfrac{1}{2}$ (see for instance Fig.\,\ref{fig:cifar}). The present work therefore fills an important gap in the literature in providing rates \eqref{eq:scaling_mm} which accurately capture the learning curves of datasets satisfying source and capacity conditions.} Further discussions of \cite{Steinwart2008SupportVM}, along with \cite{Steinwart2007FastRF,Audibert2007FastLR,Boucheron2005TheoryOC} for completeness, are provided in \ref{App:loose}. Also note that while \cite{Vecchia2021RegularizedEO} report $\sfrac{\alpha}{(1+\alpha)}$ rates under Gaussianity assumptions, they rely on very stringent assumptions which are too strong and unfulfilled in our setting. 
}

\section{Ridge classification}
\label{sec:ell2}

\subsection{Self-consistent equations}
Another standard classification method is the ridge classifier, which corresponds to minimizing
\begin{equation}
\label{eq:ridge_risk}
    \hat{\mathcal{R}}_n(w)=\frac{1}{n}\sum\limits_{\mu=1}^n\left(y^\mu -w^{\top}\psi(x^\mu)\right)^2+\lambda w^{\top}w.
\end{equation}
As previously discussed in section \ref{sec:setting}, we consider a decaying regularization $\lambda=n^{-\ell}$. The self-consistent equations characterizing the quantities $(q, m)$, read for the ridge risk \eqref{eq:ridge_risk}
\begin{align}
\label{eq:g3m_ridge}
\begin{cases}
  &\rho=\sum\limits_{k=1}^\infty\theta_k^{\star 2}\eig_k,\\
  & m=\sqrt{\frac{2}{\pi\rho}}\frac{n}{z}\sum\limits_{k=1}^\infty\frac{\eig_k^2\theta_k^{\star 2}}{1+\frac{n}{z}\eig_k},\\
  &q=\frac{n^2}{z^2}\sum\limits_{k=1}^\infty \frac{\frac{2}{\pi\rho}\theta_k^{\star 2}\eig_k^3+\frac{1+q-2m\sqrt{\frac{2}{\pi\rho}}}{n}\eig_k^2}{\left(1+\frac{n}{z}\eig_k\right)^2},\\
  &z=n\lambda+\frac{z}{n}\sum\limits_{k=1}^\infty \frac{\lambda_k}{\lambda_k+\frac{z}{n}}.
  \end{cases}
\end{align}
Further details on the derivation of \eqref{eq:g3m_ridge} from \cite{Loureiro2021CapturingTL} are provided in \ref{App:ridge}. 
Like \eqref{eq:g3m_mm}, eqs. \eqref{eq:g3m_ridge} have been formally proven in the proportional $n,p\rightarrow\infty,\,n/p=\mathcal{O}(1)$ limit in \cite{Loureiro2021CapturingTL}, but are expected to hold also in the present $n\ll p=\infty$ setting \cite{Cui2021GeneralizationER,jin2021learning}. Note that comparing to \eqref{eq:g3m_mm}, eqs. \eqref{eq:g3m_ridge} correspond to a constant student/teacher norm ratio $\hat{r}_1=\sfrac{2}{(\pi\rho)}$ and to a simple $\hat{r}_2=1+q-2m\sqrt{\sfrac{2}{(\pi\rho)}}$.
$\hat{r}_2$ moreover admits a very intuitive interpretation as the prediction mean squared error (MSE) between the true label $y=\sign{}({\theta^{\star}}^{\top}\psi(x))$ and the pre-activation linear predictor $\hat{w}^{\top}\psi(x)$, i.e. $
    \hat{r}_2=\mathbb{E}_{ \psi(x)}\left(\sign{}\left({\theta^{\star}}^{\top}\psi(x)\right)-\hat{w}^{\top}\psi(x)\right)^2.
$

\subsection{Decay rates for ridge classification}

\begin{figure}[ht!]
    \centering
    \includegraphics[scale=0.39]{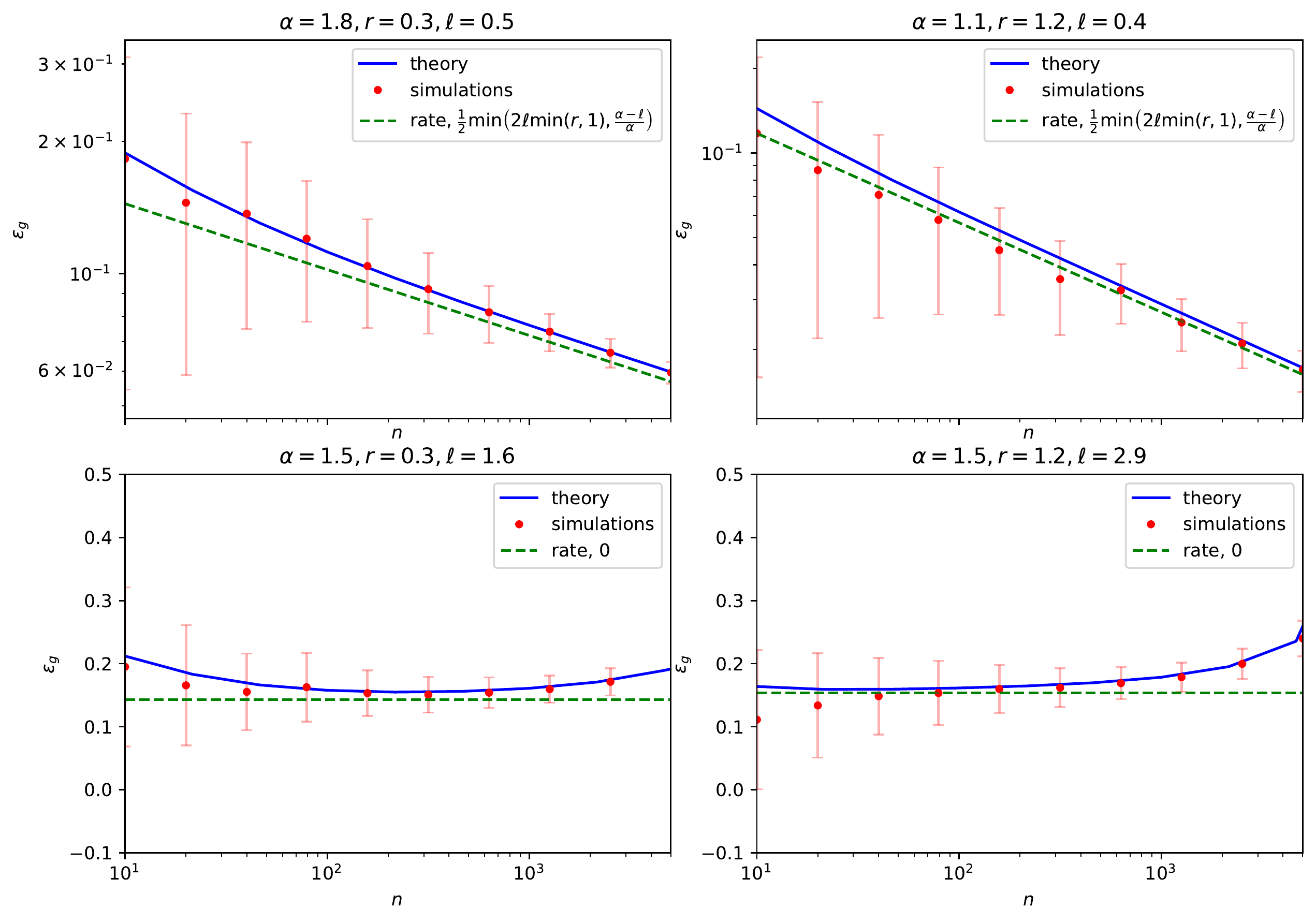}
    \caption{Misclassification error $\epsilon_g$ for ridge classification on synthetic Gaussian features, as specified in \eqref{eq:model_def}, for different source/capacity coefficients $\alpha,r$, in the effectively regularized regime $\ell\le\alpha$ (top) and unregularized regime $\ell>\alpha$ (bottom). In blue, the solution of the eqs. \eqref{eq:g3m_mm} used in the characterization \eqref{eq:error} for the misclassification error, using the \texttt{g3m} package \cite{Loureiro2021CapturingTL}. The dimension $p$ was cut-off at $10^4$. Red dots corresponds to simulations averaged over $40$ instances, for $p=10^4$. The green dashed lines indicate the power-laws \eqref{eq:ridge_exponent} (top) and \eqref{eq:plateau} (bottom)  derived in this work. The slight increase of the error for larger $n$ in the unregularized regime (bottom) is due to finite size effects of the simulations ran at $p=10^4<\infty$. Physically, it corresponds to the onset of the ascent preceding the second descent that is present for finite $p$, see \ref{App:double_descent} for further discussion. The code used for the simulations is available  \href{https://github.com/SPOC-group/Kernel_classification}{here}.}
    \label{fig:plateau}
    \label{fig:l2_error_reg}
\end{figure}

Similarly to \cite{bordelon2020,Cui2021GeneralizationER}, an analysis of the eqs.~(\ref{eq:g3m_ridge}) (see \ref{App:ridge}) reveals that, depending on how the rate of decay $\ell$ of the regularization compares to the capacity $\alpha$, two regimes (called \textit{effectively regularized} and \textit{effectively un-regularized} in \cite{Cui2021GeneralizationER} in the context of ridge regression) can be found:\\

\paragraph{Effectively regularized regime --} $\ell\le\alpha$. In this regime, an analysis of the corrections to the self-overlap $q$ and magnetization $m$, presented in \ref{App:ridge}, shows that the misclassification error scales like 
\begin{equation}
\label{eq:ridge_exponent}
    \epsilon_g\sim n^{-\frac{1}{2}\mathrm{min}\left(2\ell\mathrm{min}(r,1),\frac{\alpha-\ell}{\alpha}\right)}.
\end{equation}
 The rate \eqref{eq:ridge_exponent} compares very well to numerical simulations, see Fig.\,\ref{fig:l2_error_reg}.
 Note that the saturation for ridge happens for $r=1$, rather than $r=\sfrac{1}{2}$ as for max-margin classification (see discussion in section \ref{sec:MM}): very smooth targets $f^\star$ characterized by a source $r\ge1$ are all equally easily classified by ridge. For rougher teachers $f^\star$ characterized by $r \le 1$ however, the rate of decay of the error \eqref{eq:ridge_exponent} depends on the specific roughness of the target, even if, in contrast to max-margin, the latter belongs to $\mathcal{H}$ ($r> \sfrac{1}{2}$). Two important observations should further be made on the rates \eqref{eq:ridge_exponent}:
 \begin{itemize}
     \item If the regularization remains small (fast decay $\alpha>\ell>\sfrac{\alpha}{(1+2\alpha\mathrm{min}(r,1))}$), the decay \eqref{eq:ridge_exponent} is determined only by the data capacity $\alpha$, while the source $r$ plays no role. As a matter of fact, with insufficient regularization, the limiting factor to the learning is the tendency to overfit, which depends on the effective dimension of the data as captured by the capacity $\alpha$.
     \item For larger regularizations (slow decays $\ell<\sfrac{\alpha}{(1+2\alpha\mathrm{min}(r,1))}$), the limiting factor becomes the complexity of the teacher $\theta^\star$, as captured by the source~$r$.
 \end{itemize}
 
 \paragraph{Effectively un-regularized regime --} $\ell>\alpha$. As derived in \ref{App:ridge}, the error plateaus and stays of order $1$:
 \begin{equation}
 \label{eq:plateau}
     \epsilon_g=\mathcal{O}(1).
 \end{equation}
This plateau is further elaborated upon in \ref{App:double_descent}, and is visible in numerical experiments, see Fig.\,\ref{fig:plateau}. It corresponds to the first plateau in a double descent curve, with the second descent never happening since $p=\infty$. Intuitively, this phenomenon is attributable to the ridge classifier overfitting the labels using the small-variance directions of the data \eqref{eq:model_def}.

Interestingly, all the rates \eqref{eq:plateau} and \eqref{eq:ridge_exponent} correspond exactly (up to a factor $\sfrac{1}{2}$) to those reported in \cite{Cui2021GeneralizationER} for the MSE of ridge regression, where they are respectively called the \textit{red}, \textit{blue} and \textit{orange} exponents. Notably, the plateau \eqref{eq:plateau} at low regularizations and the $\sfrac{(\alpha-\ell)}{\alpha}$ exponent in \eqref{eq:ridge_exponent} only appeared in \cite{Cui2021GeneralizationER} for noisy cases in which the labels are corrupted by an additive noise. The fact that they hold in the present \textit{noiseless} study very temptingly suggests that model mis-specification (trying to interpolate binary labels using a linear model) effectively plays  the role of a large noise.

\subsection{Optimal rates}
\begin{figure}[ht!]
    \centering
    \includegraphics[scale=0.42]{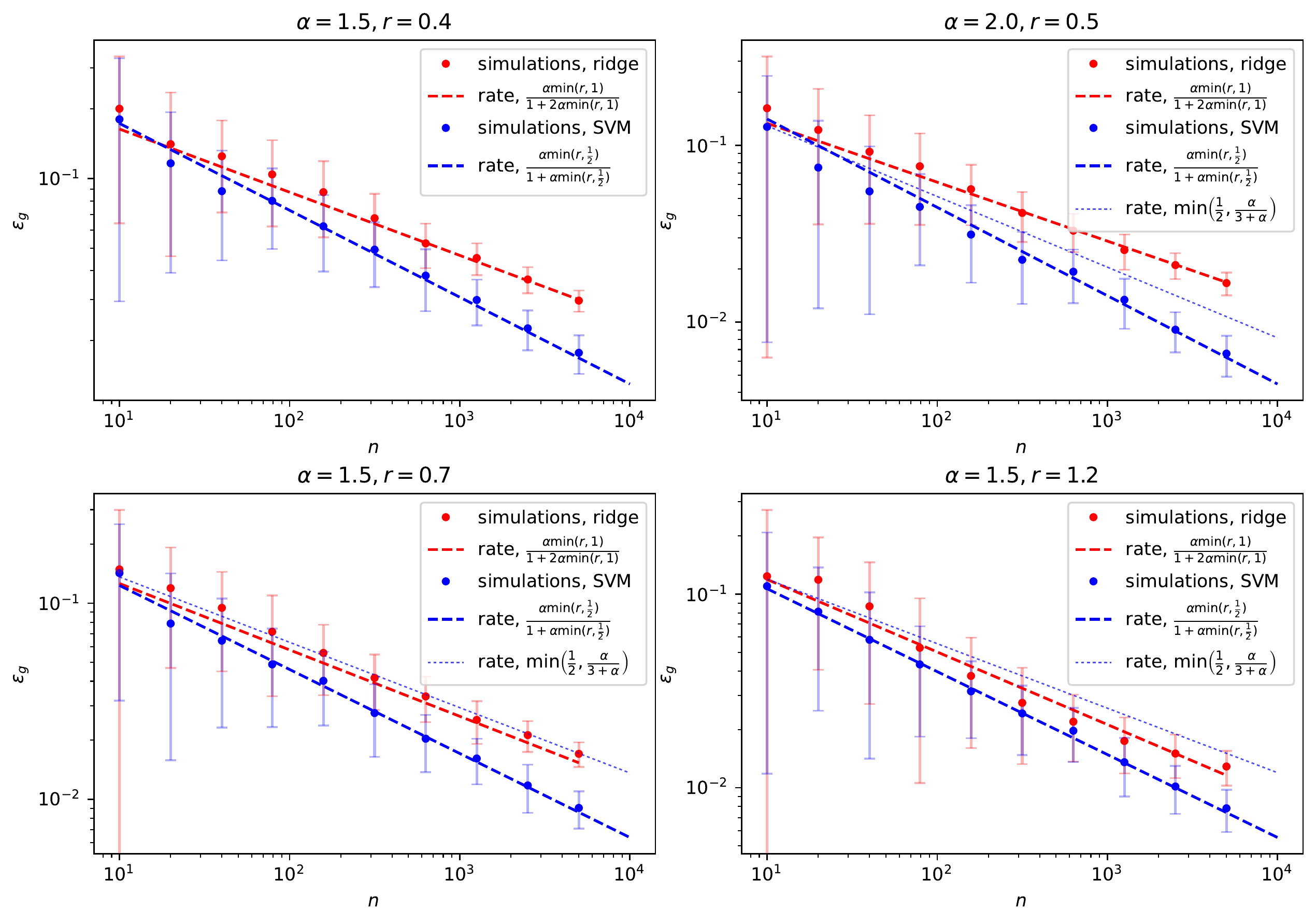}
    \caption{(red) Misclassification error $\epsilon_g$ for ridge classification on synthetic Gaussian features, as specified in \eqref{eq:model_def}, for different source/capacity coefficients $\alpha,r$, for optimal regularization $\lambda^\star$. The dimension $p$ was cut-off at $10^4$ and the regularization $\lambda$ numerically tuned to minimize the error $\epsilon_g$ for avery $n$. Red dots correspond to simulations averaged over $40$ instances, for $p=10^4$. Optimization over $\lambda$ was performed using cross validation, with the help of the python \texttt{scikit-learn GridSearchCV} package. The red dashed line represents the power-law \eqref{eq:opt_rates_ridge}. In blue, the learning curves for max-margin for the same data-set are plotted for reference, along the corresponding power law \eqref{eq:scaling_mm} (blue) and the loose classical $\min\left(\sfrac{1}{2},\sfrac{\alpha}{(3+\alpha)}\right)$ rate \cite{Steinwart2008SupportVM} (light blue), see Section\,\ref{sec:MM}. The code used for the simulations is available  \href{https://github.com/SPOC-group/Kernel_classification}{here}. }
    \label{fig:l2CV}
\end{figure}

\paragraph{Optimally regularized ridge classification}
In practice, the strength of the regularization $\lambda$ is a tunable parameter. A natural question to ask is then the one of the \textit{asymptotically optimal} regularization, that is the regularization decay rate $\ell^\star$ leading to fastest decay rates for the misclassification error. From the expressions of \eqref{eq:ridge_exponent} (which hold provided $\ell<\alpha$) and \eqref{eq:plateau} (which holds provided $\ell>\alpha$), the value of $\ell$ maximizing the error rate is found to be
\begin{equation}
\label{eq:ell_opt}
    \ell^\star=\frac{\alpha}{1+2\alpha\mathrm{min}(r,1)},
\end{equation}
and the corresponding error rate for $\epsilon_g^\star= \epsilon_g(\lambda^\star=n^{-\ell^\star})$ is 
\begin{equation}
\label{eq:opt_rates_ridge}
    \epsilon_g^\star\sim n^{-\frac{\alpha\mathrm{min}(r,1)}{1+2 \alpha\mathrm{min}(r,1)}},
\end{equation}
see the red dashed lines in Fig.\,\ref{fig:l2CV}. Coincidentally, the optimal rate \eqref{eq:opt_rates_ridge} is up to a factor $\sfrac{1}{2}$ identical to the classical optimal rate known for the rather distinct problem of the MSE of kernel ridge regression on noisy data \cite{Caponnetto2005FastRF,caponnetto2007optimal}. Like the max-margin exponent \eqref{eq:scaling_mm}, the optimal error rate for ridge \eqref{eq:opt_rates_ridge} is an increasing function of both the capacity $\alpha$ and the source $r$, i.e. of the easiness of the learning task. Note that in contrast to max-margin classification which is insensitive to the specifics of the target function $f^\star$, provided it is in $\mathcal{H}$, ridge is sensitive to the source (smoothness) $r$ of $f^\star$ up to $r=1$. \\

\paragraph{Comparison to max-margin} A comparison of the max-margin rate $a_{\mathrm{SVM}}=\sfrac{\alpha\mathrm{min}(r,\frac{1}{2})}{(1+\alpha\mathrm{min}(r,\frac{1}{2}))}$ \eqref{eq:scaling_mm} and the optimal ridge exponent $a_{r}=\sfrac{\alpha\mathrm{min}(r,1)}{(1+2\alpha\mathrm{min}(r,1))}$ \eqref{eq:opt_rates_ridge} reveals that for any $\alpha>1,r\ge 0$, $a_{\mathrm{SVM}}-a_r>0.$
In other words, the margin-maximizing SVM displays faster rates than the ridge classifier for the class of data studied \eqref{eq:model_def}, see Fig.\,\ref{fig:l2CV}.\\

We finally briefly comment on support vector proliferation. \cite{JMLR:v22:20-603,hsu2020proliferation,Ardeshir2021SupportVM} showed that in some settings almost \textit{every} training sample in $\mathcal{D}$ becomes a support vector for the SVM. In such settings, the estimators $\hat{w}$ (and hence the error $\epsilon_g$) consequently coincide for the ridge classifier and the margin-maximizing SVM. In the present setting however, the result $a_{\mathrm{SVM}}-a_r>0$ establishes that for features with a power-law decaying spectrum (\ref{eq:model_def}), there is \textit{no} such support vector proliferation. Note that this result does not follow immediately from Theorem $3$ in \cite{Ardeshir2021SupportVM}. In fact, the spiked covariance \eqref{eq:model_def}, with only a small number of important (large variance) directions and a tail of unimportant (low-variance) directions does effectively not offer enough overparametrization \cite{Bartlett2020BenignOI,hsu2020proliferation} for support vector proliferation, and the support consists only of the subset of the training set with weakest alignment with the spike.

\section{Remarks for real data-sets}
\label{sec:real}

\begin{figure}[ht!]
    \centering
    \includegraphics[scale=0.45]{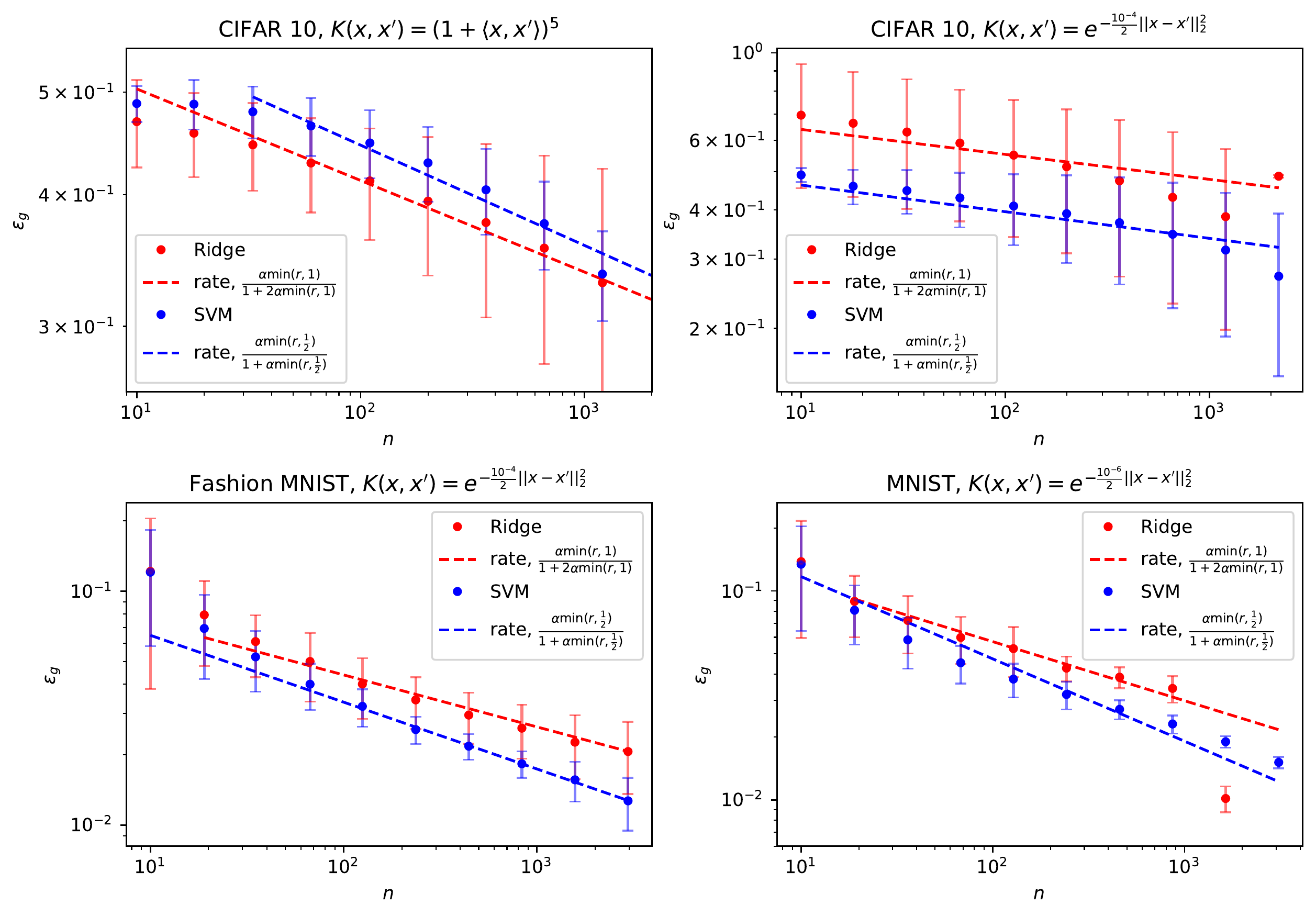}
    \caption{Dots: Misclassification error $\epsilon_g$ of kernel classification on CIFAR 10 with a polynomial kernel (top left) and an RBF kernel (top right), on Fashion MNIST with an RBF kernel (bottom left), on MNIST with an RBF kernel (bottom right), for max-margin SVM (blue) and optimally regularized ridge classification (red), using respectively the python \texttt{scikit-learn SVC} and \texttt{KernelRidge} packages. Dashed lines: Theoretical decay rates for the error $\epsilon_g$ \eqref{eq:scaling_mm} (blue) \eqref{eq:opt_rates_ridge} (red), computed from empirically estimated capacity $\alpha$ and source $r$ coefficients (see section \eqref{sec:real} and \ref{App:real} for details). \scaling{The measure coefficients are summarized in Table \ref{measured_alpha}.} The code used for the simulations is available  \href{https://github.com/SPOC-group/Kernel_classification}{here}.
    }
    \label{fig:cifar}
\end{figure}

The source and capacity condition \eqref{eq:model_def} provide a simple framework to study a large class of structured data-sets. While idealized, we observe, similarly to \cite{Cui2021GeneralizationER}, that many real data-sets seem to fall under this category of data-sets, and hence display learning curves which are to a good degree described by the rates \eqref{eq:scaling_mm} for SVM and \eqref{eq:opt_rates_ridge} for ridge classification. \hc{We present here three examples of such data-sets : a data-set of $10^4$ randomly sampled CIFAR 10 \cite{krizhevsky2009learning} images of animals (labelled $+1$) and means of transport (labelled $-1$), a data-set of $14000$ FashionMNIST \cite{xiao2017fashion} images of t-shirts (labelled $+1$) and coats (labelled $-1$), and a data-set of $14702$ MNIST \cite{lecun1998gradient} images of $8s$ (labelled $+1$) and $1s$ (labelled $-1$)}. On the one hand, the learning curves for max-margin classification and optimally regularized ridge classification were obtained using the python \texttt{scikit-learn SVC, KernelRidge} packages. On the other hand, the spectrum $\{\eig{}_k\}_k$ of the data covariance $\Sigma$ in feature space was computed, and a teacher $\theta^\star$ providing perfect classification of the data-set was fitted using margin-maximizing SVM. Then, the capacity and source coefficients $\alpha,\,r$ \eqref{eq:model_def} were estimated for the data-set by fitting $\{\eig{}_k\}_k$ and $\{\theta^\star_k\}_k$ by power laws, and the theoretical rates \eqref{eq:scaling_mm} and \eqref{eq:opt_rates_ridge} computed therefrom. More details on this method, adapted from \cite{bordelon2020,spigler2019asymptotic}, are provided in \ref{App:real}. The results of the simulations are presented in Figure\,\ref{fig:cifar} and compared to the theoretical rates \eqref{eq:scaling_mm}\eqref{eq:opt_rates_ridge} computed from the empirically evaluated source and capacity coefficients for a Radial Basis Function (RBF) kernel and a polynomial kernel of degree $5$, with overall very good agreement. \revisionL{We do not compare here with the worst case bounds because the observed values of $r<1/2$ in which case we remind the known results do not apply.}
\scaling{
\begin{table}
    \centering
\begin{tabular}{ |p{3cm}|p{3cm}||p{1.5cm}|p{1.5cm}|p{1.5cm}|p{1.5cm}|  }
 \hline
 Dataset & Kernel& $\alpha$&$r$&$a_{\mathrm{SVM}}$&$a_r$\\
 \hline
 CIFAR 10& polynomial &1.51&0.07&0.095&0.086\\
 CIFAR 10& RBF&1.005&0.07&0.067&0.063\\
 Fashion MNIST& RBF&1.72&0.23&0.28&0.22\\
MNIST& RBF&1.65&0.39&0.39&0.28\\
 \hline
\end{tabular}
\caption{\color{black}Values of the source and capacity coefficients \eqref{eq:source_capa_conditions} as estimated from the data sets, and the corresponding theoretical error rates for SVM \eqref{eq:scaling_mm} and ridge 
 \eqref{eq:opt_rates_ridge}. The details on the estimation procedure can be found in Appendix \ref{App:real}.}
\label{measured_alpha}
\end{table}
}

\section{Conclusion}

We compute the generalization error rates as a function of the source and capacity coefficients for two standard kernel classification methods, margin-maximizing SVM and ridge classification, and show that SVM classification consistently displays faster rates. Our results establish that known worst-case upper bound rates for SVM classification fail to tightly capture the rates of the class of data described by source/capacity conditions. We \scaling{illustrate empirically} that a number of real data-sets fall under this class, and display error rates which are to a very good degree described by the ones derived in this work.

\begin{acknowledgments}
We thank Loucas Pillaud-Vivien, Alessandro Rudi, Vittorio Erba and Mauro Pastore for useful discussion. We acknowledge funding from the ERC under the European Union’s Horizon 2020 Research and Innovation Programme Grant Agreement 714608-SMiLe.

\end{acknowledgments}

\clearpage

\bibliography{main}
\bibliographystyle{unsrt}

\clearpage
\appendix

\section{Rates for margin maximizing SVM}
\label{App:mm}
\label{sec:maxmargin}
In this appendix, we provide some analytical discussion of the equations \eqref{eq:g3m_mm} motivating the scaling \eqref{eq:scaling_mm} and \eqref{eq:mm_op_scaling}. We remind the risk for the hinge loss \eqref{eq:MM_risk}
\begin{equation}
    \hat{\mathcal{R}}_n(w)=\frac{1}{n}\sum\limits_{\mu=1}^n\mathrm{max}\left(0,1-y^\mu w^{\top}\psi(x)\right)+\lambda w^{\top}w,
\end{equation}
with $\lambda=0^{+}$ for max-margin. The predictor is then $\hat{y}=\mathrm{sign}(\hat{w}\cdot \psi(x))$.

\subsection{Mapping from (Loureiro et. al, 2021)}
The starting point is the closed-form asymptotic characterization of the misclassification error of \cite{Loureiro2021CapturingTL}. We begin by reviewing the main results of \cite{Loureiro2021CapturingTL}, and detail how their setting can be mapped to ours. Consider hinge regression on $n$ independent $p-$ dimensional Gaussian samples $\mathcal{D}=\{x^\mu,y^\mu\}_{\mu=1}^n$, by minimizing the empirical risk
\begin{equation}
\label{eq:app:mm:Risk_loureiro}
    \hat{\mathcal{R}}_n(w)=\sum\limits_{\mu=1}^n\mathrm{max}\left(0,1-y^\mu \frac{w^{\top}x^\mu}{\sqrt{p}}\right)+\frac{\lambda}{2} w^{\top}w
\end{equation}
for some constant regularization strength $\lambda\ge 0$. Suppose that the labels are generated from a teacher/target/oracle $\theta^\star\in\mathbb{R}^p$ as $y^\mu=\sign(\theta^{\star\top}x^\mu)$. Then provided the assumptions
\begin{itemize}
    \item \textbf{A.1} $n\gg1,\,p\gg1,\,\frac{n}{p}=\mathcal{O}(1)$,
    \item \textbf{A.2} $0<\frac{1}{p}||\theta^\star||^2_2<\infty$,
\end{itemize}
are satisfied, there exist constants $C,c,c'>0$ so that for all $0<\epsilon<c'$,
\begin{equation}
    \mathbb{P}\left(\left|\epsilon_g-\frac{1}{\pi}\mathrm{arccos}\left(\frac{m^\star}{\sqrt{\rho q^\star}}\right)\right|>\epsilon\right)<\frac{C}{\epsilon}e^{-c n \epsilon^2},
\end{equation}
where $\rho=\theta^{\star\top}\Sigma\theta^\star/p$ and $m^\star,\,q^\star$ are the solutions of the \textit{fixed point equations}
\begin{align}
\label{eq:app:mm:loureiro_g3m}
    &\begin{cases} m= \frac{\hat{m}}{p}~\mathrm{tr}\frac{\Sigma\theta^\star\theta^{\star \top}\Sigma}{\lambda+\hat{V}\Sigma}\\
    q=\frac{1}{p}\mathrm{tr}\frac{\hat{m}^2\Sigma\theta^\star\theta^{\star \top}\Sigma^2+\hat{q}\Sigma^2}{(\lambda+ \hat{V}\Sigma)^2}\\
    V=\frac{1}{p}\mathrm{tr}\frac{\Sigma}{\lambda+\hat{V}\Sigma}
    \end{cases},\nonumber\\
    &\begin{cases}
    \begin{aligned}
    \hat{m}&=\frac{\frac{n}{p}}{V\sqrt{\rho}}\frac{1}{2\pi}\Bigg\{
    \sqrt{2\pi}\left[
    \mathrm{erf}\left(\frac{1}{\sqrt{2q(1-\eta)}}\right)-\mathrm{erf}\left(\frac{1-V}{\sqrt{2q(1-\eta)}}\right)
    \right]\nonumber\\
    &+2\sqrt{q(1-\eta)}\left(e^{-\frac{1}{2q(1-\eta)}}-e^{-\frac{(1-V)^2}{2q(1-\eta)}}\right)+\sqrt{2\pi}V\left(1+\mathrm{erf}\left(\frac{1-V}{\sqrt{2q(1-\eta)}}\right)
    \right)
    \Bigg\}\end{aligned}\\
    \hat{q}=\frac{n}{p}\Bigg[ \int\limits_{-\infty}^\frac{1-V}{\sqrt{q}}dx\frac{e^{-\frac{1}{2}x^2}}{\sqrt{2\pi}}\left[1+\mathrm{erf}(\sqrt{\frac{\eta}{2(1-\eta)}}x)
     \right]
     +\frac{1}{V^2}\int\limits_{\frac{1-V}{\sqrt{q}}}^{\frac{1}{\sqrt{q}}}dx\frac{e^{-\frac{1}{2}x^2}}{\sqrt{2\pi}}\left[1+\mathrm{erf}(\sqrt{\frac{\eta}{2(1-\eta)}}x)
     \right](1-\sqrt{q}x)^2\Bigg]
     \\
    \hat{V}=\frac{\frac{n}{p}}{V}\int\limits_{\frac{1-V}{\sqrt{q}}}^{\frac{1}{\sqrt{q}}}dx\frac{e^{-\frac{1}{2}x^2}}{\sqrt{2\pi}} \left[1+\mathrm{erf}(\sqrt{\frac{\eta}{2(1-\eta)}}x)
     \right]\end{cases},
\end{align}
with $\eta=m^2/(\rho q)$, and $\Sigma$ the covariance of the samples $x$. The limit $\lambda=0^+$ can be taken using the rescaling \cite{aubin2020generalization}
\begin{align}
    &\hat{m}\leftarrow \frac{\hat{m}}{\lambda},
    && \hat{q}\leftarrow \frac{\hat{q}}{\lambda},
    && \hat{V}\leftarrow \frac{\hat{V}}{\lambda},
    && V\leftarrow \lambda V.
\end{align}
The equations \eqref{eq:app:mm:loureiro_g3m} simplify in this limit to 
\begin{align}
\label{eq:app:mm:loureiro_g3m_rescaled}
    &\begin{cases} m= \frac{\hat{m}}{p}~\mathrm{tr}\frac{\Sigma\theta^\star\theta^{\star \top}\Sigma}{1+\hat{V}\Sigma}\\
    q=\frac{1}{p}\mathrm{tr}\frac{\hat{m}^2\Sigma\theta^\star\theta^{\star \top}\Sigma^2+\hat{q}\Sigma^2}{(1+ \hat{V}\Sigma)^2}\\
    V=\frac{1}{p}\mathrm{tr}\frac{\Sigma}{1+\hat{V}\Sigma}
    \end{cases},\nonumber\\
    &\begin{cases}
    \hat{m}=\frac{\frac{n}{p}}{V\sqrt{\rho}}\frac{1}{2\pi} 
    \left(
    \sqrt{2\pi}(1+\mathrm{erf}(\frac{1}{\sqrt{2q(1-\eta)}}))+2
    e^{-\frac{1}{2q(1-\eta)}}\sqrt{q(1-\eta)}
    \right)\\
    \hat{q}=\frac{\frac{n}{p}}{V^2}\int\limits_{-\infty}^{\frac{1}{\sqrt{q}}}dx\frac{e^{-\frac{1}{2}x^2}}{\sqrt{2\pi}} \left[1+\mathrm{erf}(\sqrt{\frac{\eta}{2(1-\eta)}}x)
     \right](1-\sqrt{q}x)^2\\
     \hat{V}=\frac{\frac{n}{p}}{V}\int\limits_{-\infty}^{\frac{1}{\sqrt{q}}}dx\frac{e^{-\frac{1}{2}x^2}}{\sqrt{2\pi}} \left[1+\mathrm{erf}(\sqrt{\frac{\eta}{2(1-\eta)}}x)
     \right]\end{cases}.
\end{align}

Note that the risk studied by \cite{Loureiro2021CapturingTL} \eqref{eq:app:mm:Risk_loureiro} differs from the one we consider \eqref{eq:MM_risk} by the scaling $1/\sqrt{p}$, the missing $1/n$ in front of the sum, and a factor $2$ for the regularization strength. All those scalings can be absorbed in $\lambda\leftarrow 2\lambda/n$ and $\Sigma\leftarrow \Sigma/p$, leading to \eqref{eq:g3m_mm}:

\begin{align}
    \begin{cases} m=\hat{m} p~\mathrm{tr}\frac{\Sigma\theta^\star\theta^{\star \top}\Sigma}{1+p\hat{V}\Sigma}\\
    q=p\mathrm{tr}\frac{p\hat{m}^2\Sigma\theta^\star\theta^{\star \top}\Sigma^2+\hat{q}\Sigma^2}{(1+p \hat{V}\Sigma)^2}\\
    V=\mathrm{tr}\frac{\Sigma}{1+p\hat{V}\Sigma}\\
    \end{cases},
    &&\begin{cases}\hat{m}=\frac{\frac{n}{p}}{V\sqrt{\rho}}\frac{1}{2\pi} 
    \left(
    \sqrt{2\pi}(1+\mathrm{erf}(\frac{1}{\sqrt{2q(1-\eta)}}))+2
    e^{-\frac{1}{2q(1-\eta)}}\sqrt{q(1-\eta)}
    \right)\\
    \hat{q}=\frac{\frac{n}{p}}{V^2}\int\limits_{-\infty}^{\frac{1}{\sqrt{q}}}dx\frac{e^{-\frac{1}{2}x^2}}{\sqrt{2\pi}} \left[1+\mathrm{erf}(\sqrt{\frac{\eta}{2(1-\eta)}}x)
     \right](1-\sqrt{q}x)^2\\
     \hat{V}=\frac{\frac{n}{p}}{V}\int\limits_{-\infty}^{\frac{1}{\sqrt{q}}}dx\frac{e^{-\frac{1}{2}x^2}}{\sqrt{2\pi}} \left[1+\mathrm{erf}(\sqrt{\frac{\eta}{2(1-\eta)}}x)
     \right]\end{cases},
\end{align}

thereby completeing the mapping from the setup in \cite{Loureiro2021CapturingTL} to the present setting. 

\subsection{Equations for max-margin under source and capacity conditions}

In the following, we detail the asymptotic scaling analysis of the equations \eqref{eq:g3m_mm}. For a diagonal covariance $\Sigma=\mathrm{diag}(\omega_1,...,\omega_p)$ and $\theta^\star=(\theta_k^\star)_k$, \eqref{eq:g3m_mm} reads

\begin{align}
\label{eq:app:mm:g3m_with_components}
    \begin{cases}
    \rho=\sum\limits_{k=1}^p\theta_k^{\star 2}\eig_k\\
    m=\hat{m}p\sum\limits_{k=1}^d\frac{\eig_k^2\theta_k^{\star 2}}{1+\hat{V}p\eig_k}\\
    q=p\sum\limits_{k=1}^p\frac{p\hat{m}^2\theta_k^{\star 2}\eig_k^3+\hat{q}\eig_k^2}{(1+\hat{V}p\eig_k)^2}\\
    V=\sum\limits_{k=1}^p\frac{\eig_k}{1+\hat{V}p\eig_k}\\
    \end{cases},
    &&\begin{cases}\hat{m}=\frac{\frac{n}{p}}{V\sqrt{\rho}}\frac{1}{2\pi} 
    \left(
    \sqrt{2\pi}(1+\mathrm{erf}(\frac{1}{\sqrt{2q(1-\eta)}}))+2
    e^{-\frac{1}{2q(1-\eta)}}\sqrt{q(1-\eta)}
    \right)\\
    \hat{q}=\frac{\frac{n}{p}}{V^2}\int\limits_{-\infty}^{\frac{1}{\sqrt{q}}}dx\frac{e^{-\frac{1}{2}x^2}}{\sqrt{2\pi}} \left[1+\mathrm{erf}(\sqrt{\frac{\eta}{2(1-\eta)}}x)
     \right](1-\sqrt{q}x)^2\\
     \hat{V}=\frac{\frac{n}{p}}{V}\int\limits_{-\infty}^{\frac{1}{\sqrt{q}}}dx\frac{e^{-\frac{1}{2}x^2}}{\sqrt{2\pi}} \left[1+\mathrm{erf}(\sqrt{\frac{\eta}{2(1-\eta)}}x)
     \right]\end{cases}.
\end{align}

To simplify the equations \eqref{eq:app:mm:g3m_with_components}, we introduce the auxiliary variables $z=\frac{n}{p}/\hat{V}$, $\hat{r}_1=\hat{m}/\hat{V}$ and $\hat{r}_2=(n\hat{q})/(p\hat{V}^2)$. The intuitive meaning of $z$ is loosely that of an effective regularization. In the context of kernel ridge \textit{regression} where a similar variable appears, the role of $z$ as an effective regularizing term is quite clear in \cite{Cui2021GeneralizationER}. We also refer the reader to the discussion in \cite{Loureiro2021CapturingTL}, also for ridge regression, where the role of $\hat{V}$ as parametrizing the gap between training and test error is mentionned. $\hat{r}_1$ is to be regarded as the ratio between the norm of the estimator $\hat{w}$ minimizing the risk \eqref{eq:MM_risk}, and the norm of the teacher $\theta^\star$. Introducing these variables in \eqref{eq:app:mm:g3m_with_components} allows to have a well defined $p=\infty$ limit, which reads

\begin{align}
\label{eq:app:mm:g3m_final}
    \begin{cases}
    \rho=\sum\limits_{k=1}^\infty\theta_k^{\star 2}\eig_k\\
    m=\hat{r}_1\frac{n}{z}\sum\limits_{k=1}^\infty\frac{\eig_k^2\theta_k^{\star 2}}{1+\frac{n}{z}\eig_k}\\
    q=\frac{n^2}{z^2}\sum\limits_{k=1}^\infty\frac{\hat{r}_1^2\theta_k^{\star 2}\eig_k^3+\frac{1}{n}\hat{r}_2\eig_k^2}{(1+\frac{n}{z}\eig_k)^2}\\
    V=\sum\limits_{k=1}^d\frac{\eig_k}{1+\frac{n}{z}\eig_k}\\
    \end{cases},
    &&\begin{cases}\hat{r}_1=\frac{\hat{m}}{\hat{V}}=\frac{1}{2\pi\sqrt{\rho}} 
    \frac{\left(
    \sqrt{2\pi}(1+\mathrm{erf}(\frac{1}{\sqrt{2q(1-\eta)}}))+2
    e^{-\frac{1}{2q(1-\eta)}}\sqrt{q(1-\eta)}
    \right)}{\int\limits_{-\infty}^{\frac{1}{\sqrt{q}}}dx\frac{e^{-\frac{1}{2}x^2}}{\sqrt{2\pi}} \left[1+\mathrm{erf}(\sqrt{\frac{\eta}{2(1-\eta)}}x)
     \right]}\\
    \hat{r}_2=\frac{\frac{n}{p}\hat{q}}{\hat{V}^2}=\frac{\int\limits_{-\infty}^{\frac{1}{\sqrt{q}}}dx\frac{e^{-\frac{1}{2}x^2}}{\sqrt{2\pi}} \left[1+\mathrm{erf}(\sqrt{\frac{\eta}{2(1-\eta)}}x)
     \right](1-\sqrt{q}x)^2}{\left(\int\limits_{-\infty}^{\frac{1}{\sqrt{q}}}dx\frac{e^{-\frac{1}{2}x^2}}{\sqrt{2\pi}} \left[1+\mathrm{erf}(\sqrt{\frac{\eta}{2(1-\eta)}}x)
     \right]\right)^2}\\
    \end{cases}.
\end{align}

From \eqref{eq:app:mm:g3m_with_components}, $z$ satisfies the self-consistent equation
\begin{equation}
    z=\frac{\frac{z}{n}\sum\limits_{k=1}^\infty \frac{\eig_k}{\frac{z}{n}+\eig_k}}{\int\limits_{-\infty}^{\frac{1}{\sqrt{q}}}dx\frac{e^{-\frac{1}{2}x^2}}{\sqrt{2\pi}} \left[1+\mathrm{erf}(\sqrt{\frac{\eta}{2(1-\eta)}}x)
     \right]}\approx
    \frac{\left(\frac{z}{n}\right)^{1-\frac{1}{\alpha}}\int_{\left(\frac{z}{n}\right)^{\frac{1}{\alpha}}}^\infty \frac{dx}{1+x^\alpha}}{\int\limits_{-\infty}^{\frac{1}{\sqrt{q}}}dx\frac{e^{-\frac{1}{2}x^2}}{\sqrt{2\pi}} \left[1+\mathrm{erf}(\sqrt{\frac{\eta}{2(1-\eta)}}x) \right]},
\end{equation}
where a Riemann approximation of the sum was used. We introduce its to-be-determined scaling $\gamma$,
\begin{equation}
\label{eq:app:mm:intro_gamma}
    z\sim C_z n^{1-\alpha\gamma},
\end{equation}
where $C_z$ designates the prefactor.
In the following, we discuss the scaling and the corrections of the order parameters and express them as a function of $\gamma$, before determining its value using the numerical solution of \eqref{eq:app:mm:g3m_final}. Note that the scaling \eqref{eq:app:mm:intro_gamma} implies in particular the following scaling for the integral 
\begin{equation}
\label{eq:app:mm:scaling_integral}
    \int\limits_{-\infty}^{\frac{1}{\sqrt{q}}}dx\frac{e^{-\frac{1}{2}x^2}}{\sqrt{2\pi}} \left[1+\mathrm{erf}(\sqrt{\frac{\eta}{2(1-\eta)}}x)\right]\sim n^{\gamma-1}.
\end{equation}

\subsection{First order corrections}

We now plug the source and capacity ansatz \eqref{eq:source_capa_conditions} into the equations \eqref{eq:app:mm:g3m_final}
\begin{align}
    \eig_k\sim k^{-\alpha},&&\theta_k^\star\sim k^{-\frac{1+\alpha(2r-1)}{2}},
\end{align}
yielding
\begin{align}
    \begin{cases}
    \rho=&\sum\limits_{k=1}^\infty k^{-1-2\alpha r}\\
    m=&\hat{r}_1\sum\limits_{k=1}^\infty\frac{k^{-1-2\alpha r}}{1+\frac{z}{n}k^\alpha}\\
    q=&\hat{r}_1^2\sum\limits_{k=1}^\infty\frac{k^{-1-2\alpha r}}{(1+\frac{z}{n}k^\alpha)^2}\nonumber\\
    &+\frac{\hat{r}_2}{n}\sum\limits_{k=1}^\infty\frac{1}{(1+\frac{z}{n}k^\alpha)^2}\\
    V=&\frac{z}{n}\sum\limits_{k=1}^\infty\frac{1}{1+\frac{z}{n}k^\alpha}
    \end{cases},
    &&\begin{cases}\hat{r}_1=\frac{1}{2\pi\sqrt{\rho}} 
    \frac{\left(
    \sqrt{2\pi}(1+\mathrm{erf}(\frac{1}{\sqrt{2q(1-\eta)}}))+2
    e^{-\frac{1}{2q(1-\eta)}}\sqrt{q(1-\eta)}
    \right)}{\int\limits_{-\infty}^{\frac{1}{\sqrt{q}}}dx\frac{e^{-\frac{1}{2}x^2}}{\sqrt{2\pi}} \left[1+\mathrm{erf}(\sqrt{\frac{\eta}{2(1-\eta)}}x)
     \right]}\\
    \hat{r}_2=\frac{\int\limits_{-\infty}^{\frac{1}{\sqrt{q}}}dx\frac{e^{-\frac{1}{2}x^2}}{\sqrt{2\pi}} \left[1+\mathrm{erf}(\sqrt{\frac{\eta}{2(1-\eta)}}x)
     \right](1-\sqrt{q}x)^2}{\left(\int\limits_{-\infty}^{\frac{1}{\sqrt{q}}}dx\frac{e^{-\frac{1}{2}x^2}}{\sqrt{2\pi}} \left[1+\mathrm{erf}(\sqrt{\frac{\eta}{2(1-\eta)}}x)
     \right]\right)^2}\\
    \end{cases}.
\end{align}
Massaging the expression for $m$ (and remembering $z\sim n^{1-\alpha\gamma}$ \eqref{eq:app:mm:intro_gamma})

\begin{align}
\label{eq:app:mm:expansion_m}
    m&=\hat{r}_1\rho-\hat{r}_1\sum\limits_{k=1}^\infty\frac{k^{-1-2\alpha r}C_z\left(\frac{k}{n^\gamma}\right)^\alpha}{1+C_z\left(\frac{k}{n^\gamma}\right)^\alpha}\nonumber\\
    &=\hat{r}_1\rho-C_z\hat{r}_1n^{-\gamma\alpha}\sum\limits_{k=1}^\infty\frac{k^{-1-2\alpha (r-\frac{1}{2})}}{1+C_z\left(\frac{k}{n^\gamma}\right)^\alpha}\nonumber\\
    &\approx \hat{r}_1\rho- \mathbbm{1}_{r\ge\frac{1}{2}}\hat{r}_1C_z\left[
    n^{-\gamma\alpha}\sum\limits_{k=1}^{n^\gamma}\frac{k^{-1-2\alpha (r-\frac{1}{2})}}{1+C_z\left(\frac{k}{n^\gamma}\right)^\alpha}+n^{-2\gamma\alpha r}\int\limits_1^\infty \frac{x^{-1+\alpha(1-2r)}}{1+C_zx^\alpha}
    \right]\nonumber\\
    &~~~~-\mathbbm{1}_{r\le\frac{1}{2}}C_z\hat{r}_1n^{-2\gamma\alpha r }\int\limits_0^\infty \frac{x^{-1+\alpha(1-2r)}}{1+C_zx^\alpha}\nonumber\\
    &= \hat{r}_1\left[\rho-\mathbbm{1}_{r\ge\frac{1}{2}}C_zA^n_mn^{-\gamma\alpha}-C_z n^{-2\gamma\alpha r }\int\limits_{\Theta(r-\frac{1}{2})}^\infty \frac{x^{-1+\alpha(1-2r)}}{1+C_zx^\alpha}+o\left(\mathrm{max}\left(n^{-\gamma\alpha},n^{-2\gamma\alpha r}\right)\right)\right],
\end{align}
where we used the shorthand
\begin{equation}
    A_m^n\overset{\mathrm{def}}{=}\sum\limits_{k=1}^{n^\gamma}\frac{k^{-1-2\alpha (r-\frac{1}{2})}}{1+C_z\left(\frac{k}{n^\gamma}\right)^\alpha}.
\end{equation}

Massaging by the same token the equation  for $q$,
\begin{align}
\label{eq:app:mm:expansion_q}
    q&\approx\hat{r}_1^2\rho - \hat{r}_1^2\sum\limits_{k=1}^\infty\frac{k^{-1-2\alpha r}\left(C_z^2\left(\frac{k}{n^\gamma}\right)^{2\alpha}+2C_z\left(\frac{k}{n^\gamma}\right)^\alpha\right)}{\left(1+C_z\left(\frac{k}{n^\gamma}\right)^\alpha\right)^2}+\hat{r}_2n^{\gamma-1}\int\limits_0^\infty \frac{1}{(1+C_zx^\alpha)^2}\nonumber\\
    &\approx\hat{r}_1^2\rho+\hat{r}_2n^{\gamma-1}\int\limits_0^\infty \frac{1}{(1+C_zx^\alpha)^2}\nonumber\\
    &~~~~-2\hat{r}_1^2C_z\Bigg[
    \mathbbm{1}_{r\ge\frac{1}{2}}\left(
   n^{-\gamma\alpha}\sum\limits_{k=1}^{n^\gamma}\frac{k^{-1-2\alpha (r-\frac{1}{2})}}{\left(1+C_z\left(\frac{k}{n^\gamma}\right)^\alpha\right)^2}+n^{-2\gamma\alpha r}\int\limits_1^\infty \frac{x^{-1+\alpha(1-2r)}}{(1+C_zx^\alpha)^2}
    \right)\nonumber\\
    &~~~~+\mathbbm{1}_{r\le\frac{1}{2}}n^{-2\gamma\alpha r }\int\limits_0^\infty \frac{x^{-1+\alpha(1-2r)}}{(1+C_zx^\alpha)^2}
    \Bigg]\nonumber\\
    &~~~~ -\hat{r}_1^2C_z^2\Bigg[
    \mathbbm{1}_{r\ge1}\Bigg(
    n^{-2\gamma\alpha}\sum\limits_{k=1}^{n^\gamma}\frac{k^{-1-2\alpha (r-1)}}{\left(1+C_z\left(\frac{k}{n^\gamma}\right)^\alpha\right)^2}\nonumber\\
    &~~~~+n^{-2\gamma\alpha r}\int\limits_1^\infty \frac{x^{-1+\alpha(2-2r)}}{(1+C_zx^\alpha)^2}
    \Bigg)+\mathbbm{1}_{r\le1}n^{-2\gamma\alpha r }\int\limits_0^\infty \frac{x^{-1+\alpha(2-2r)}}{(1+C_zx^\alpha)^2}
    \Bigg]\nonumber\\
    &=\hat{r}_1^2\Bigg[\rho+\frac{\hat{r}_2}{\hat{r}_1^2}n^{\gamma-1}\int\limits_0^\infty \frac{1}{(1+C_zx^\alpha)^2}-\mathbbm{1}_{r\ge\frac{1}{2}}2C_zA^n_q n^{-\gamma\alpha}-\mathbbm{1}_{r\ge 1}C_z^2B^n_qn^{-2\gamma\alpha}\nonumber\\
    &~~~~-n^{-2\gamma\alpha r}\Bigg(2C_z\int\limits_{\Theta(r-\frac{1}{2})}^\infty \frac{x^{-1+\alpha(1-2r)}}{(1+C_zx^\alpha)^2}+C_z^2\int\limits_{\Theta(r-1)}^\infty \frac{x^{-1+\alpha(2-2r)}}{(1+C_zx^\alpha)^2}\Bigg)\nonumber\\
    &~~~~+o\left(\mathrm{max}\left(\frac{\hat{r}_2}{\hat{r}_1^2}n^{\gamma-1},n^{-\gamma\alpha},n^{-2\gamma\alpha r}\right)\right)\Bigg],
\end{align}
where we defined
\begin{align}
    A_q^n\overset{\mathrm{def}}{=}\sum\limits_{k=1}^{n^\gamma}\frac{k^{-1-2\alpha (r-\frac{1}{2})}}{\left(1+C_z\left(\frac{k}{n^\gamma}\right)^\alpha\right)^2}, && B^n_q\overset{\mathrm{def}}{=}\sum\limits_{k=1}^{n^\gamma}\frac{k^{-1-2\alpha (r-1)}}{\left(1+C_z\left(\frac{k}{n^\gamma}\right)^\alpha\right)^2}.
\end{align}


An important remark on \eqref{eq:app:mm:expansion_m} and \eqref{eq:app:mm:expansion_q} is that $A^m_q$ and $A^n_m$ tend to the same limit as $n\rightarrow \infty$,
\begin{equation}
    A^\infty_{m,q}=\sum\limits_{k=1}^\infty k^{-1-2\alpha (r-\frac{1}{2})},
\end{equation}
with (for $r\ge \frac{1}{2}$ since there is an indicator function in front of $A^n_{m,q}$)
\begin{align}
    \begin{cases}
    n^{-\gamma\alpha}|A^n_m-A^\infty|=&\frac{n^{-2\alpha\gamma r}}{2\alpha(r-\frac{1}{2})}+\mathbbm{1}_{\frac{1}{2}\le r\le1}C_zn^{-2\alpha\gamma r}\int\limits_0^1\frac{x^{-1-2\alpha(r-1)}}{1+C_zx^\alpha}\nonumber\\
    &+\mathbbm{1}_{r\ge 1}C_zn^{-2\gamma\alpha}\sum\limits_{k=1}^\infty k^{-1-2\alpha(r-1)},\\
    n^{-\gamma\alpha}|A^n_q-A^\infty|=&\frac{n^{-2\alpha\gamma r}}{2\alpha(r-\frac{1}{2})}+2\mathbbm{1}_{\frac{1}{2}\le r\le1}C_zn^{-2\alpha\gamma r}\int\limits_0^1\frac{x^{-1-2\alpha(r-1)}}{(1+C_zx^\alpha)^2}\nonumber\\
    &+2\mathbbm{1}_{r\ge 1}C_zn^{-2\gamma\alpha}\sum\limits_{k=1}^\infty k^{-1-2\alpha(r-1)}+o(n^{-2\gamma\alpha}).
    \end{cases}
\end{align}
More straigtforwardly, 
\begin{equation}
    B^\infty_q=\sum\limits_{k=1}^\infty k^{-1-2\alpha (r-1 )}.
\end{equation}

Finally, remark that the expansions \eqref{eq:app:mm:expansion_m} and \eqref{eq:app:mm:expansion_q} imply that $m\sim \hat{r}_1\rho$ and $q\sim\hat{r}_1^2\rho$, which echoes the intuitive meaning of $m,q$ as $\langle \hat{w},\theta^\star\rangle_{L^2(\mathcal{X})},\, ||\hat{w}||^2_{L^2(\mathcal{X})}$ and of $\hat{r}_1$ as $||\hat{w}||^2_{L^2(\mathcal{X})}/||\theta^\star||^2_{L^2(\mathcal{X})}$, see also discussion in section \ref{sec:setting} under equation \eqref{eq:m_q_physical_meaning}.\\

The expansions \eqref{eq:app:mm:expansion_m} and \eqref{eq:app:mm:expansion_q} can be plugged into the expression of the cosine similarity $\eta$ \eqref{eq:error} to access the decay rate for the misclassification error.
\begin{align}
\label{eq:app:mm:expansion_eta}
   \eta&=\Bigg(1-\mathbbm{1}_{r\ge\frac{1}{2}}C_z\frac{A^n_m}{\rho}n^{-\gamma\alpha}-n^{-2\gamma\alpha r }\frac{C_z}{\rho}\int\limits_{\Theta(r-\frac{1}{2})}^\infty \frac{x^{-1+\alpha(1-2r)}}{1+C_zx^\alpha}\Bigg)^2\Bigg/\Bigg(1+\nonumber\\
   ~~&\frac{\hat{r}_2}{\rho\hat{r}_1^2}n^{\gamma-1}\int\limits_0^\infty \frac{1}{(1+x^\alpha)^2}-\frac{\mathbbm{1}_{r\ge\frac{1}{2}}2C_zA^n_qn^{-\gamma\alpha}-\mathbbm{1}_{r\ge 1}C_z^2B^n_qn^{-2\gamma\alpha}}{\rho}\nonumber\\
   ~~& -n^{-2\gamma\alpha r}\frac{C_z}{\rho}\Bigg(2\int\limits_{\Theta(r-\frac{1}{2})}^\infty \frac{x^{-1+\alpha(1-2r)}}{(1+C_zx^\alpha)^2}+C_z\int\limits_{\Theta(r-1)}^\infty \frac{x^{-1+\alpha(2-2r)}}{(1+C_zx^\alpha)^2}\Bigg)\Bigg)^{-1}\nonumber\\
   &=1-\mathcal{O}\left(\frac{\hat{r}_2}{\rho\hat{r}_1^2}n^{\gamma-1}\right)+\mathbbm{1}_{r\ge 1}\mathcal{O}\left(n^{-2\gamma\alpha}\right)+\mathcal{O}\left(n^{-2\gamma\alpha r}\right),
\end{align}
where we used $A^\infty_q=A^\infty_m$, meaning the leading order of the $n^{-\alpha\gamma}$ term cancels out. The scaling of the other order parameters $q, m, \hat{r}_1, \hat{r}_2$ can at this point also be deduced. To see how, notice that
\begin{align}
    \int\limits_{-\infty}^{\frac{1}{\sqrt{q}}}dx\frac{e^{-\frac{1}{2}x^2}}{\sqrt{2\pi}} \left[1+\mathrm{erf}(\sqrt{\frac{\eta}{2(1-\eta)}}x)\right]\approx \frac{1}{\sqrt{q}}
    \int\limits_{-\infty}^{1}dx\frac{e^{-\frac{x^2}{2q}}}{\sqrt{2\pi}} \left[1+\mathrm{erf}(\sqrt{\frac{\eta}{2q(1-\eta)}}x)\right]\sim \frac{1}{\sqrt{q}},
\end{align}
which together with the observation \eqref{eq:app:mm:scaling_integral} implies
\begin{equation}
    q\sim \hat{r}_1^2\sim n^{2(1-\gamma)}.
\end{equation}
In parallel, it follows from a similar change of variables in the expression of $\hat{r}_2$ that
\begin{equation}
     \hat{r}_2\sim \sqrt{q}\sim\hat{r}_1\sim n^{1-\gamma} .
\end{equation}
Then, returning to \eqref{eq:app:mm:expansion_eta},
\begin{equation}
\label{eq:app:mm:expansion_eta_final}
    1-\eta=\mathcal{O}\left(n^{2(\gamma-1)}\right)+\mathbbm{1}_{r\ge 1}\mathcal{O}\left(n^{-2\gamma\alpha}\right)+\mathcal{O}\left(n^{-2\gamma\alpha r}\right).
\end{equation}

The final step is to determine the value of $\gamma$. Suppose self-consistently that the leading rate in \eqref{eq:app:mm:expansion_q} is $2(\gamma-1)$. Since from the physical meaning of the order parameters \eqref{eq:m_q_physical_meaning} one expects the leading corrections of $m$ and $q$ to share the same decay rate \scaling{-- since in \eqref{eq:m_q_physical_meaning}, only $\hat{w}$ admits a correction with $n$ --}, and since the leading corrections of $m$ has rate $-2\gamma\alpha \mathrm{min}(r,1/2)$ \eqref{eq:app:mm:expansion_m}, it follows from equating the two rates that $\gamma$ should be

\begin{equation}
\label{eq:app:gamma}
    \gamma=\frac{1}{1+\alpha\mathrm{min}(r,\frac{1}{2})}.
\end{equation}
Plugging the expression \eqref{eq:app:gamma} back in \eqref{eq:app:mm:expansion_q}, it can be checked that the $2(\gamma-1)$ term is indeed leading. \eqref{eq:app:gamma} compares very well with numerical simulations, and with the numerical solution of \eqref{eq:app:mm:g3m_final}, see Fig.\,\ref{fig:mm_error}. 

Assuming \eqref{eq:app:gamma} to be true, since $\epsilon_g=\mathrm{cos}^{-1}(\eta)\sim\sqrt{1-\eta}$,
\begin{equation}
    \epsilon_g\sim n^{-\gamma\alpha\mathrm{min}(r,\frac{1}{2})}=n^{-\frac{\alpha\mathrm{min}(r,\frac{1}{2})}{1+\alpha\mathrm{min}(r,\frac{1}{2})}},
\end{equation}
which is the error rate for max-margin classification \eqref{eq:scaling_mm}.

\section{Rates for regularized hinge classification}
\label{App:reg}
In this Appendix, we show that margin-maximizing SVM is optimal for classifying teachers with source $r\le \frac{1}{2}$, i.e. in $L^2(\mathcal{X})\setminus \mathcal{H}$, and argue why this is also expected to extend to teachers in $\mathcal{H}$. In other words, we give evidence that the optimal regularization in the hinge classification risk \eqref{eq:MM_risk} on the dataset \eqref{eq:model_def} is for vanishing $\lambda=0$. To this end, we derive the rates for the misclassification error of ERM on \eqref{eq:MM_risk} for generic regularization $\lambda$, building upon the results of \cite{Loureiro2021CapturingTL} in similar fashion to Appendix \ref{App:mm}.\\

\subsection{Regularized hinge classification under source and capacity conditions}
We remind the fixed point equations for the risk \eqref{eq:MM_risk} (see Appendix \ref{App:mm}):
\begin{align}
    &\begin{cases} 
    m=\hat{m} p~\mathrm{tr}\frac{\Sigma\theta^\star\theta^{\star \top}\Sigma}{\frac{n\lambda}{2}+p\hat{V}\Sigma}\\
    q=p~\mathrm{tr}\frac{p\hat{m}^2\Sigma\theta^\star\theta^{\star \top}\Sigma^2+\hat{q}\Sigma^2}{(\frac{n\lambda}{2}+p \hat{V}\Sigma)^2}\\
    V=\mathrm{tr}\frac{\Sigma}{\frac{n\lambda}{2}+p\hat{V}\Sigma}\\
    \end{cases},\nonumber\\
    &\begin{cases}
    \begin{aligned}
    \hat{m}&=\frac{\frac{n}{p}}{V\sqrt{\rho}}\frac{1}{2\pi}\Bigg\{
    \sqrt{2\pi}\left[
    \mathrm{erf}\left(\frac{1}{\sqrt{2q(1-\eta)}}\right)-\mathrm{erf}\left(\frac{1-V}{\sqrt{2q(1-\eta)}}\right)
    \right]\nonumber\\
    &+2\sqrt{q(1-\eta)}\left(e^{-\frac{1}{2q(1-\eta)}}-e^{-\frac{(1-V)^2}{2q(1-\eta)}}\right)+\sqrt{2\pi}V\left(1+\mathrm{erf}\left(\frac{1-V}{\sqrt{2q(1-\eta)}}\right)
    \right)
    \Bigg\}\end{aligned}\\
    \hat{q}=\frac{n}{p}\Bigg[ \int\limits_{-\infty}^\frac{1-V}{\sqrt{q}}dx\frac{e^{-\frac{1}{2}x^2}}{\sqrt{2\pi}}\left[1+\mathrm{erf}(\sqrt{\frac{\eta}{2(1-\eta)}}x)
     \right]
     +\frac{1}{V^2}\int\limits_{\frac{1-V}{\sqrt{q}}}^{\frac{1}{\sqrt{q}}}dx\frac{e^{-\frac{1}{2}x^2}}{\sqrt{2\pi}}\left[1+\mathrm{erf}(\sqrt{\frac{\eta}{2(1-\eta)}}x)
     \right](1-\sqrt{q}x)^2\Bigg]
     \\
    \hat{V}=\frac{\frac{n}{p}}{V}\int\limits_{\frac{1-V}{\sqrt{q}}}^{\frac{1}{\sqrt{q}}}dx\frac{e^{-\frac{1}{2}x^2}}{\sqrt{2\pi}} \left[1+\mathrm{erf}(\sqrt{\frac{\eta}{2(1-\eta)}}x)
     \right]\end{cases}.\\
\label{eq:app:reg:loureiro_g3m_reg}
\end{align}

Introducing the effective regularization $z=\frac{\alpha n\lambda}{2\hat{V}}$, satisfying

\begin{equation}
    z=\frac{\frac{z}{n}\sum\limits_{k=1}^\infty \frac{\lambda_k}{\frac{z}{n}+\lambda_k}}{\int\limits_{\frac{1-V}{\sqrt{q}}}^{\frac{1}{\sqrt{q}}}dx\frac{e^{-\frac{1}{2}x^2}}{\sqrt{2\pi}} \left[1+\mathrm{erf}(\sqrt{\frac{\eta}{2(1-\eta)}}x)
     \right]}\approx
    \frac{\left(\frac{z}{n}\right)^{1-\frac{1}{\alpha}}\int_{\left(\frac{z}{n}\right)^{\frac{1}{\alpha}}}^\infty \frac{dx}{1+x^\alpha}}{\int\limits_{\frac{1-V}{\sqrt{q}}}^{\frac{1}{\sqrt{q}}}dx\frac{e^{-\frac{1}{2}x^2}}{\sqrt{2\pi}} \left[1+\mathrm{erf}(\sqrt{\frac{\eta}{2(1-\eta)}}x) \right]},
\label{eq:app:reg:z_self}
\end{equation}
the equations \eqref{eq:app:reg:loureiro_g3m_reg} can be rewritten as
\begin{align}
    \begin{cases}
    \rho=\sum\limits_{k=1}^\infty\theta_k^2\lambda_k\\
    m=\hat{r}_1\frac{n}{z}\sum\limits_{k=1}^\infty\frac{\lambda_k^2\theta_k^2}{1+\frac{n}{z}\lambda_k}\\
    q=\frac{n^2}{z^2}\sum\limits_{k=1}^\infty\frac{\hat{r}_1^2\theta_k^2\lambda_k^3+\frac{1}{n}\hat{r}_2\lambda_k^2}{(1+\frac{n}{z}\lambda_k)^2}\\
    V=\frac{1}{n\lambda}\sum\limits_{k=1}^d\frac{\lambda_k}{1+\frac{n}{z}\lambda_k}\\
    \end{cases}
\end{align}
\begin{align}
    \begin{cases}\hat{r}_1=&\frac{\hat{m}}{\hat{V}}=\frac{1}{2\pi\sqrt{\rho}} 
    \frac{
    \sqrt{2\pi}\left[
    \mathrm{erf}\left(\frac{1}{\sqrt{2q(1-\eta)}}\right)-\mathrm{erf}\left(\frac{1-V}{\sqrt{2q(1-\eta)}}\right)
    \right]
    }{\int\limits_{\frac{1-V}{\sqrt{q}}}^{\frac{1}{\sqrt{q}}}dx\frac{e^{-\frac{1}{2}x^2}}{\sqrt{2\pi}} \left[1+\mathrm{erf}(\sqrt{\frac{\eta}{2(1-\eta)}}x)
     \right]}\nonumber \\
     &+\frac{1}{2\pi\sqrt{\rho}} 
    \frac{
    2\sqrt{q(1-\eta)}\left(e^{-\frac{1}{2q(1-\eta)}}-e^{-\frac{(1-V)^2}{2q(1-\eta)}}\right)+\sqrt{2\pi}V\left(1+\mathrm{erf}\left(\frac{1-V}{\sqrt{2q(1-\eta)}}\right)
    \right)
    }{\int\limits_{\frac{1-V}{\sqrt{q}}}^{\frac{1}{\sqrt{q}}}dx\frac{e^{-\frac{1}{2}x^2}}{\sqrt{2\pi}} \left[1+\mathrm{erf}(\sqrt{\frac{\eta}{2(1-\eta)}}x)
     \right]}\\
    \hat{r}_2=&\frac{\alpha\hat{q}}{\hat{V}^2}=\frac{V^2\int\limits_{-\infty}^\frac{1-V}{\sqrt{q}}dx\frac{e^{-\frac{1}{2}x^2}}{\sqrt{2\pi}}\left[1+\mathrm{erf}(\sqrt{\frac{\eta}{2(1-\eta)}}x)
     \right]
     +\int\limits_{\frac{1-V}{\sqrt{q}}}^{\frac{1}{\sqrt{q}}}dx\frac{e^{-\frac{1}{2}x^2}}{\sqrt{2\pi}}\left[1+\mathrm{erf}(\sqrt{\frac{\eta}{2(1-\eta)}}x)
     \right](1-\sqrt{q}x)^2}{\left(\int\limits_{\frac{1-V}{\sqrt{q}}}^{\frac{1}{\sqrt{q}}}dx\frac{e^{-\frac{1}{2}x^2}}{\sqrt{2\pi}} \left[1+\mathrm{erf}(\sqrt{\frac{\eta}{2(1-\eta)}}x)
     \right]\right)^2}\\
    \end{cases}.
\end{align}

Specializing to the source/capacity power-law forms \eqref{eq:model_def}
\begin{align}
    \eig_k\sim k^{-\alpha},&&\theta_k\sim k^{-\frac{1+\alpha(2r-1)}{2}},
\end{align}
one finally reaches
\begin{align}
    \begin{cases}
    \rho=\sum\limits_{k=1}^\infty k^{-1-2\alpha r}\\
    m=\hat{r}_1\sum\limits_{k=1}^\infty\frac{k^{-1-2\alpha r}}{1+\frac{z}{n}k^\alpha}\\
    q=\hat{r}_1^2\sum\limits_{k=1}^\infty\frac{k^{-1-2\alpha r}}{(1+\frac{z}{n}k^\alpha)^2}+\frac{\hat{r}_2}{n}\sum\limits_{k=1}^\infty\frac{1}{(1+\frac{z}{n}k^\alpha)^2}\\
    V=\frac{1}{n\lambda}\frac{z}{n}\sum\limits_{k=1}^\infty\frac{1}{1+\frac{z}{n}k^\alpha}
    \end{cases}
\end{align}
\begin{align}
    \begin{cases}\hat{r}_1=&\frac{\hat{m}}{\hat{V}}=\frac{1}{2\pi\sqrt{\rho}} 
    \frac{
    \sqrt{2\pi}\left[
    \mathrm{erf}\left(\frac{1}{\sqrt{2q(1-\eta)}}\right)-\mathrm{erf}\left(\frac{1-V}{\sqrt{2q(1-\eta)}}\right)
    \right]
    }{\int\limits_{\frac{1-V}{\sqrt{q}}}^{\frac{1}{\sqrt{q}}}dx\frac{e^{-\frac{1}{2}x^2}}{\sqrt{2\pi}} \left[1+\mathrm{erf}(\sqrt{\frac{\eta}{2(1-\eta)}}x)
     \right]}\\
     &+\frac{1}{2\pi\sqrt{\rho}} 
    \frac{
    2\sqrt{q(1-\eta)}\left(e^{-\frac{1}{2q(1-\eta)}}-e^{-\frac{(1-V)^2}{2q(1-\eta)}}\right)+\sqrt{2\pi}V\left(1+\mathrm{erf}\left(\frac{1-V}{\sqrt{2q(1-\eta)}}\right)
    \right)
    }{\int\limits_{\frac{1-V}{\sqrt{q}}}^{\frac{1}{\sqrt{q}}}dx\frac{e^{-\frac{1}{2}x^2}}{\sqrt{2\pi}} \left[1+\mathrm{erf}(\sqrt{\frac{\eta}{2(1-\eta)}}x)
     \right]}\\
    \hat{r}_2=&\frac{\alpha\hat{q}}{\hat{V}^2}=\frac{V^2\int\limits_{-\infty}^\frac{1-V}{\sqrt{q}}dx\frac{e^{-\frac{1}{2}x^2}}{\sqrt{2\pi}}\left[1+\mathrm{erf}(\sqrt{\frac{\eta}{2(1-\eta)}}x)
     \right]
     +\int\limits_{\frac{1-V}{\sqrt{q}}}^{\frac{1}{\sqrt{q}}}dx\frac{e^{-\frac{1}{2}x^2}}{\sqrt{2\pi}}\left[1+\mathrm{erf}(\sqrt{\frac{\eta}{2(1-\eta)}}x)
     \right](1-\sqrt{q}x)^2}{\left(\int\limits_{\frac{1-V}{\sqrt{q}}}^{\frac{1}{\sqrt{q}}}dx\frac{e^{-\frac{1}{2}x^2}}{\sqrt{2\pi}} \left[1+\mathrm{erf}(\sqrt{\frac{\eta}{2(1-\eta)}}x)
     \right]\right)^2}\\
    \end{cases}
\label{eq:app:reg:g3m_final}
\end{align}

\subsection{Rate analysis for targets outside the Hilbert space}
In the following, we deliver an asymptotic analysis of eqs. \eqref{eq:app:reg:g3m_final}. To that end, we first ascertain the scaling of the effective regularization $z$ using the self-consistent equation \eqref{eq:app:reg:z_self}. Depending on the scaling of the order parameter $V$, two regimes can be distinguished.

\paragraph{Effectively regularized regime $V\xrightarrow[]{n\rightarrow\infty}\infty$} In this regime the denominator of \eqref{eq:app:reg:z_self} scales like
\begin{align}
    \int\limits_{\frac{1-V}{\sqrt{q}}}^{\frac{1}{\sqrt{q}}}dx\frac{e^{-\frac{1}{2}x^2}}{\sqrt{2\pi}} \left[1+\mathrm{erf}(\sqrt{\frac{\eta}{2(1-\eta)}}x) \right]&\approx\frac{1}{\sqrt{q}}\int\limits_{-\infty}^{1}dx\frac{e^{-\frac{1}{2q}x^2}}{\sqrt{2\pi}} \left[1+\mathrm{erf}(\sqrt{\frac{\eta}{2q(1-\eta)}}x) \right]=\mathcal{O}\left(\frac{1}{\sqrt{q}}\right) ,
\label{eq:reg:deno_z1}
\end{align}
which is exactly the scaling for the corresponding integral in the vanishing regularization case $\lambda=0^+$, see Appendix \ref{App:mm} eq. \eqref{eq:app:mm:scaling_integral}. In this regime, all the discussion in Appendix \ref{App:mm} then carries through and the error rate coincides with the one for margin-maximizing SVM \eqref{eq:scaling_mm}
\begin{equation}
\label{eq:app:reg:error_unreg_regime}
    \epsilon_g\sim n^{-\frac{\alpha\mathrm{min}(r,\frac{1}{2})}{1+\alpha\mathrm{min}(r,\frac{1}{2})}}.
\end{equation}

\paragraph{Effectively regularized regime $V\xrightarrow[]{n\rightarrow\infty}0$}
The integral in the denominator in \eqref{eq:app:reg:z_self} then admits the following scaling
\begin{align}
    \int\limits_{\frac{1-V}{\sqrt{q}}}^{\frac{1}{\sqrt{q}}}dx\frac{e^{-\frac{1}{2}x^2}}{\sqrt{2\pi}} \left[1+\mathrm{erf}(\sqrt{\frac{\eta}{2(1-\eta)}}x) \right]&=\frac{1}{\sqrt{q}}\int\limits_{1-V}^{1}dx\frac{e^{-\frac{1}{2q}x^2}}{\sqrt{2\pi}} \left[1+\mathrm{erf}(\sqrt{\frac{\eta}{2q(1-\eta)}}x) \right]=\mathcal{O}\left(\frac{V}{\sqrt{q}}\right).
    \label{eq:reg:deno_z}
\end{align}
This means in particular that
\begin{equation}
\label{eq:app:reg:scaling_z}
    z=\frac{Vn\lambda}{\int\limits_{\frac{1-V}{\sqrt{q}}}^{\frac{1}{\sqrt{q}}}dx\frac{e^{-\frac{1}{2}x^2}}{\sqrt{2\pi}} \left[1+\mathrm{erf}(\sqrt{\frac{\eta}{2(1-\eta)}}x) \right]}=\mathcal{O}(n\lambda\sqrt{q})\overset{\mathrm{def}}{\sim}C_z n^{1-\alpha\gamma},
\end{equation}
where we introduced similarly to the max-margin case the to-be determined parameter $\gamma$, see also Appendix \ref{App:mm}. Note that it follows from equation \eqref{eq:app:reg:scaling_z} and the definition of $\lambda=n^{- \ell}$ that 
\begin{equation}
  q\sim n^{2(\ell-\alpha\gamma)}.  
\end{equation}
Since $n\lambda V\sim (z/n)^{1-\frac{1}{\alpha}}$, one also has that
\begin{equation}
\label{eq:app:reg:V}
    V\sim n^{-1+\gamma(1-\alpha)+\ell}.
\end{equation}
In particular, it can be seen from \eqref{eq:app:reg:V} that the assumption $V\xrightarrow[]{n\rightarrow\infty}0$ only holds for large enough regularizations (slow enough decays $\ell<\ell^\star$ for some limiting value $\ell^\star$), satisfying
\begin{equation}
\label{eq:app:reg:condition_l_star}
    -1+\gamma(1-\alpha)+\ell<0.
\end{equation}
The regularization decay $\ell=\ell^\star$ gives the boundary between the effectively regularized and unregularized regime. Because of this, the rate of $q$ at $\ell^\star$ should coincide with its max-margin rate \eqref{eq:mm_op_scaling}. This, together with \eqref{eq:app:reg:condition_l_star}, allow to determine $\ell^\star$ as the solution of the system (denoting $\gamma^\star$ the value of $\gamma$ at $\ell=\ell^\star$)
\begin{align}
    \begin{cases}
     & -1+\gamma^\star(1-\alpha)+\ell=0,\\
     & \ell^\star-\alpha\gamma^\star=\frac{\alpha\mathrm{min}(r,\frac{1}{2})}{1+\alpha\mathrm{min}(r,\frac{1}{2})},
    \end{cases}
\end{align}
viz.
\begin{equation}
    \ell^\star=\alpha \frac{1+\mathrm{min}(r,\frac{1}{2})}{1+\alpha\mathrm{min}(r,\frac{1}{2})}.
\end{equation}
Summarizing, for $\ell>\ell^\star$, in the effectively unregularized regime, the max-margin scalings \eqref{eq:mm_op_scaling}, \eqref{eq:scaling_mm} hold. In the following, we focus on pursuing the discussion for the new $\ell<\ell^\star$ (effectively regularized) regime.

\begin{figure}[ht]
    \centering
    \includegraphics[scale=0.46]{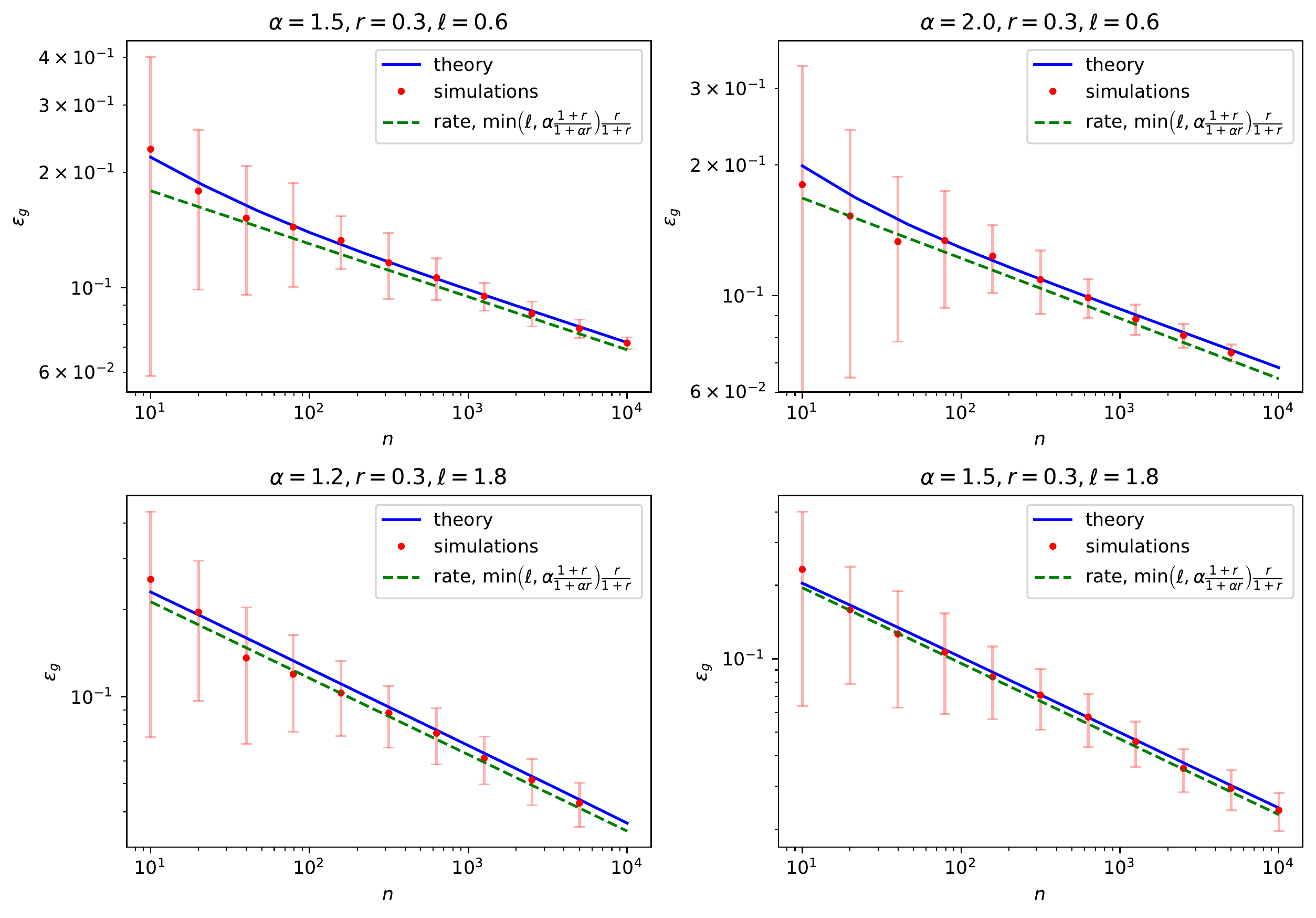}
    \caption{Misclassification error $\epsilon_g$ for hinge classification on synthetic Gaussian data, as specified in \eqref{eq:model_def}, for different source/capacity coefficients $\alpha,r$, for a regularization $\lambda=n^{-\ell}$. In blue, the solution of the closed set of equations \eqref{eq:app:reg:loureiro_g3m_reg} used in the characterization \eqref{eq:error} for the misclassification error, using the \texttt{g3m} package \cite{Loureiro2021CapturingTL}. The dimension $p$ was cut-off at $10^4$. Red dots corresponds to simulations using the \texttt{scikit-learn SVC} package and averaged over $40$ instances, for $p=10^4$. The green dashed line indicates the power-law rate \eqref{eq:app:reg:error_decay_compact} derived in this work. }
    \label{fig:app:reg:noiselessreg}
\end{figure}

The expansion for $m$ and $q$ carry over in similar fashion to max-margin (see Appendix \ref{App:mm}), yielding
\begin{equation}
    \begin{cases}
     m=\hat{r}_1\left[\rho-\mathbbm{1}_{r\ge\frac{1}{2}}C_zA^n_mn^{-\gamma\alpha}-C_zn^{-2\gamma\alpha r }\int\limits_{\Theta(r-\frac{1}{2})}^\infty \frac{x^{-1+\alpha(1-2r)}}{1+C_zx^\alpha}\right]+o\left(\mathrm{max}\left(n^{-\gamma\alpha},n^{-2\gamma\alpha r}\right)\right),\\
     \begin{aligned}
     q&=\hat{r}_1^2\Bigg[\rho+\frac{\hat{r}_2}{\hat{r}_1^2}n^{\gamma-1}\int\limits_0^\infty \frac{1}{(1+C_zx^\alpha)^2}-\mathbbm{1}_{r\ge\frac{1}{2}}2C_zA^n_q n^{-\gamma\alpha}-\mathbbm{1}_{r\ge 1}C_z^2B^n_qn^{-2\gamma\alpha}\nonumber\\
    &~~-n^{-2\gamma\alpha r}\Bigg(2C_z\int\limits_{\Theta(r-\frac{1}{2})}^\infty \frac{x^{-1+\alpha(1-2r)}}{(1+C_zx^\alpha)^2}+C_z^2\int\limits_{\Theta(r-1)}^\infty \frac{x^{-1+\alpha(2-2r)}}{(1+C_zx^\alpha)^2}\Bigg)\Bigg]\nonumber\\
    &~~+o\left(\mathrm{max}\left(\frac{\hat{r}_2}{\hat{r}_1^2}n^{\gamma-1},n^{-\gamma\alpha},n^{-2\gamma\alpha r}\right)\right).
    \end{aligned}
    \end{cases}
\end{equation}
Therefore $\hat{r}_1\sim \sqrt{q}\sim n^{\ell-\alpha\gamma}$, while a rescaling of the integrals in the equations for $\hat{r}_2$ also reveal that $\hat{r}_2\sim\hat{r}_1\sim \sqrt{q}\sim n^{\ell-\alpha\gamma}$. Summarizing
\begin{equation}
    m\sim \hat{r}_2\sim\hat{r}_1\sim \sqrt{q}\sim n^{\ell-\alpha\gamma}.
\end{equation}
Note that the mutual scaling of the order parameters $m, q, \hat{r}_1$ is the same as for max-margin classification \eqref{eq:mm_op_scaling}, which is the mutual scaling directly following from the physical interpretation of $m,q$ as $\langle \hat{w},\theta^\star\rangle_{L^2(\mathcal{X})},\, ||\hat{w}||^2_{L^2(\mathcal{X})}$ and of $\hat{r}_1$ as $||\hat{w}||^2_{L^2(\mathcal{X})}/||\theta^\star||^2_{L^2(\mathcal{X})}$, see also discussion in section \ref{sec:setting} under equation \eqref{eq:m_q_physical_meaning}.
Finally, 

\begin{align}
\label{eq:app:reg:expansion_eta}
   \eta&=\Bigg(1-\mathbbm{1}_{r\ge\frac{1}{2}}C_z\frac{A^n_m}{\rho}n^{-\gamma\alpha}-n^{-2\gamma\alpha r }\frac{C_z}{\rho}\int\limits_{\Theta(r-\frac{1}{2})}^\infty \frac{x^{-1+\alpha(1-2r)}}{1+C_zx^\alpha}\Bigg)^2\Bigg/\Bigg( 1+\nonumber\\
   &~~\frac{\hat{r}_2}{\rho\hat{r}_1^2}n^{\gamma-1}\int\limits_0^\infty \frac{1}{(1+x^\alpha)^2}-\frac{\mathbbm{1}_{r\ge\frac{1}{2}}2C_zA^n_qn^{-\gamma\alpha}-\mathbbm{1}_{r\ge 1}C_z^2B^n_qn^{-2\gamma\alpha}}{\rho}\nonumber\\
   &~~-n^{-2\gamma\alpha r}\frac{C_z}{\rho}\Bigg(2\int\limits_{\Theta(r-\frac{1}{2})}^\infty \frac{x^{-1+\alpha(1-2r)}}{(1+C_zx^\alpha)^2}+C_z\int\limits_{\Theta(r-1)}^\infty \frac{x^{-1+\alpha(2-2r)}}{(1+C_zx^\alpha)^2}\Bigg)\Bigg)^{-1}\nonumber\\
   &=1-\mathcal{O}\left(\frac{\hat{r}_2}{\rho\hat{r}_1^2}n^{\gamma-1}\right)+\mathbbm{1}_{r\ge 1}\mathcal{O}\left(n^{-2\gamma\alpha}\right)+\mathcal{O}\left(n^{-2\gamma\alpha r}\right)\nonumber\\
   &=1-\mathcal{O}\left(n^{\gamma(\alpha+1)-1-\ell}\right)+\mathcal{O}\left(n^{-2\alpha\gamma r}\right),
\end{align}
where we specialized to $r\le \frac{1}{2}$ in the last line, thereby focusing on target functions $f^\star\in L^2(\mathcal{X})\setminus\mathcal{H}$. At this point, it remains to heuristically determine $\gamma$, a rigorous analytical derivation of our results being left for future work. Making the two (numerically verified) assumptions
\begin{assumption}
Like the max-margin case, $1- \eta\sim q^{-1}$,
\end{assumption}
\begin{assumption}
The term of rate $2\alpha\gamma r$ dominates in \eqref{eq:app:reg:expansion_eta}.
\end{assumption}
$\gamma$ can be guessed as
\begin{equation}
    \gamma=\frac{\ell}{\alpha\left(1+r\right)}.
\end{equation}
Note that consistently, the value of $\gamma$ at the boundary $\ell=\ell^\star$ with the unregularized regime $\gamma^\star$ coincides with its max-margin value \eqref{eq:app:gamma}. In the effectively regularized regime, under these assumptions, we thus conjecture the error to scale for $r\le\frac{1}{2}$ as
\begin{equation}
\label{eq:app:reg:error_reg_regime}
    \epsilon_g\sim \sqrt{1-\eta}\sim n^{\alpha\gamma-\ell}\sim n^{-\ell\frac{r}{1+r}}.
\end{equation}

Finally observe that the rates for the effectively unregularized regime $\ell\ge\ell^\star$ \eqref{eq:app:reg:error_unreg_regime} and regularized regime $\ell\le\ell^\star$ \eqref{eq:app:reg:error_reg_regime} can be subsumed in the more compact form, still for $r\le\frac{1}{2}$:
\begin{equation}
    \label{eq:app:reg:error_decay_compact}
    \epsilon_g\sim n^{-\mathrm{min}(\ell,\ell^\star)\frac{r}{1+r}}= n^{-\mathrm{min}\left(\ell,\alpha\frac{1+ r}{1+\alpha r}\right)\frac{r}{1+r}}.
\end{equation}
Fig.\,\ref{fig:app:reg:noiselessreg} contrasts the rates \eqref{eq:app:reg:error_decay_compact} to the numerical solution of the equations \eqref{eq:app:reg:loureiro_g3m_reg} and to numerical simulations, and displays a very good agreement.
The main conclusion from \eqref{eq:app:reg:error_decay_compact} is that for any $\ell$, the rate is \textit{necessarily slower} than $\ell^\star r/(1+r)=\alpha r/(1+\alpha r)$, which is the max-margin rate \eqref{eq:scaling_mm}. Therefore, the max-margin rate \eqref{eq:scaling_mm} is optimal for $r\le \frac{1}{2}$, and is achieved for any regularization $\lambda$ decaying at least as fast as $n^{-\ell^\star}$. In particular, it is achieved for the limit case $\ell=\infty$, i.e. $\lambda=0^+$ a.k.a. the max-margin case, thereby suggesting that no regularization is optimal for rough enough target functions $f^\star\in L^2(\mathcal{X})\setminus \mathcal{H}$. Since regularization is not needed for hard teachers (small source $r$), we do not a fortiori expect it to help for easier, smoother teachers $f^\star\in\mathcal{H}$ characterized by source $r\ge \frac{1}{2}$. This suggests that $\lambda=0^+$ (max-margin) should be optimal for all sources $r$, i.e. any target $f^\star\in L^2(\mathcal{X})$. While this conjecture is observed to hold in numerical simulations, a more thorough theoretical analysis of the error rates for $r\ge\frac{1}{2}$ is nevertheless warranted. We leave this more challenging analysis to future work.

\section{Rates for ridge classification}
\label{App:ridge}
In this section we derive the rates \eqref{eq:ridge_exponent} for ridge classifiers. 
\subsection{Ridge classification under the source and capacity conditions}
The equations on $m,\,q$ for ridge classification \cite{Loureiro2021CapturingTL} can as in Appendix \ref{App:mm} be adapted to the $p=\infty,\,n\gg1$ limit as
\begin{align}
\label{eq:app:ridge:g3m_generic}
    \begin{cases} m=\hat{m} p~\mathrm{tr}\frac{\Sigma\theta^\star\theta^{\star \top}\Sigma}{n\lambda+p\hat{V}\Sigma}\\
    q=p~\mathrm{tr}\frac{p\hat{m}^2\Sigma\theta^\star\theta^{\star \top}\Sigma^2+\hat{q}\Sigma^2}{(n\lambda+p \hat{V}\Sigma)^2}\\
    V=\mathrm{tr}\frac{\Sigma}{n\lambda+p\hat{V}\Sigma}\\
    \end{cases},
    &&\begin{cases}
    &\hat{V}=\frac{\frac{n}{p}}{1+V}\\
    &\hat{q}=\frac{n}{p}\frac{1+q-2m\sqrt{\frac{2}{\pi\rho}}}{(1+V)^2}=\frac{\hat{V}^2}{\frac{n}{p}}\left(1+q-2m\sqrt{\frac{2}{\pi\rho}}\right)\\
    &\hat{m}=\sqrt{\frac{2}{\pi\rho}}\frac{\frac{n}{p}}{1+V}=\sqrt{\frac{2}{\pi\rho}}\hat{V}.
    \end{cases}
\end{align}

Following \cite{Cui2021GeneralizationER}, we introduce the effective regularization $z$
\begin{align}
\label{eq:app:ridge:z}
    z=\frac{\frac{n}{p} n\lambda}{\hat{V}}=n\lambda+\frac{z}{n}\sum\limits_{k=1}^\infty \frac{\lambda_k}{\lambda_k+\frac{z}{n}}\approx n\lambda+\left(\frac{z}{n}\right)^{1-\frac{1}{\alpha}}\int\limits_{\left(\frac{z}{n}\right)^{\frac{1}{\alpha}}}^\infty \frac{dx}{1+x^\alpha}.
\end{align}
 Like in Appendix \ref{App:mm}, we denote $z\sim C_z n^{1-\alpha \gamma}$ the scaling of $z$. In contrast to Appendices \ref{App:mm} and \ref{App:reg} for SVM, the exponent $\gamma$ for ridge classifier can be straightforwardly determined. Depending on the relative strength of the two terms in \eqref{eq:app:ridge:z}, two possible values for $\gamma$ exist:
\begin{itemize}
    \item $\gamma=\frac{\ell}{\alpha}$ if $n\lambda\gg (z/n)^{1-1/\alpha}$ in \eqref{eq:app:ridge:z}, i.e. $\ell<\alpha$. This correspond to the \textit{effectively regularized regime} \cite{Cui2021GeneralizationER}.
    \item $\gamma=1$ if $n\lambda\ll (z/n)^{1-1/\alpha}$ in \eqref{eq:app:ridge:z}, i.e. $\ell>\alpha$. In this regime, the regularization $\lambda$ is negligible and therefore the learning is \textit{effectively un-regularized} \cite{Cui2021GeneralizationER}.
\end{itemize}

Rewriting the equations \eqref{eq:app:ridge:g3m_generic} using \eqref{eq:app:ridge:z}, the equations \eqref{eq:g3m_ridge} of the main text are reached (see section \ref{sec:ell2}):
\begin{align}
  \begin{cases}
  &\rho=\sum\limits_{k=1}^\infty k^{-1-2\alpha r}\\
  & m=\sqrt{\frac{2}{\pi\rho}}\sum\limits_{k=1}^\infty \frac{k^{-1-2\alpha r}}{1+\frac{z}{n}k^\alpha}\\
  &q=\frac{2}{\pi\rho}\sum\limits_{k=1}^\infty \frac{k^{-1-2\alpha r}}{\left(1+\frac{z}{n}k^\alpha\right)^2}+\frac{1+q-2m\sqrt{\frac{2}{\pi\rho}}}{n}\sum\limits_{k=1}^\infty \frac{1}{\left(1+\frac{z}{n}k^\alpha\right)^2}\\
  & V=\frac{z}{n^2\lambda}\sum\limits_{k=1}^\infty \frac{1}{1+\frac{z}{n}k^\alpha}
  \end{cases}.
\end{align}

\subsection{Scaling analysis}
We proceed to deliver an scaling analysis of eqs. \eqref{eq:g3m_ridge}. We know from Appendix \ref{sec:maxmargin} the following scalings

\begin{equation}
    \begin{cases}
     m=\sqrt{\frac{2}{\pi\rho}}\left[\rho-\mathbbm{1}_{r\ge\frac{1}{2}}C_zA^n_mn^{-\gamma\alpha}-n^{-2\gamma\alpha r }C_z\int\limits_{\Theta(r-\frac{1}{2})}^\infty \frac{x^{-1+\alpha(1-2r)}}{1+C_zx^\alpha}\right]+o\left(\mathrm{max}\left(n^{-\gamma\alpha},n^{-2\gamma\alpha r}\right)\right),\\
     \begin{aligned}
     q&=\frac{2}{\pi\rho}\Bigg[\rho+\frac{\pi\rho C_z}{2}\left(1+q-2m\sqrt{\frac{2}{\pi\rho}}\right)n^{\gamma-1}\int\limits_0^\infty \frac{1}{(1+C_zx^\alpha)^2}-\mathbbm{1}_{r\ge\frac{1}{2}}2C_zA^n_q n^{-\gamma\alpha}-\mathbbm{1}_{r\ge 1}C_z^2B^n_qn^{-2\gamma\alpha}\nonumber\\
    &~~-n^{-2\gamma\alpha r}\Bigg(2C_z\int\limits_{\Theta(r-\frac{1}{2})}^\infty \frac{x^{-1+\alpha(1-2r)}}{(1+C_zx^\alpha)^2}+C_z^2\int\limits_{\Theta(r-1)}^\infty \frac{x^{-1+\alpha(2-2r)}}{(1+C_zx^\alpha)^2}\Bigg)\Bigg]\nonumber\\
    &~~+o\left(\mathrm{max}\left(\frac{\hat{r}_2}{\hat{r}_1^2}n^{\gamma-1},n^{-\gamma\alpha},n^{-2\gamma\alpha r}\right)\right).
    \end{aligned}
    \end{cases}
\end{equation}
Once again, we defined the shorthands
\begin{align}
    A_m^n\overset{\mathrm{def}}{=}\sum\limits_{k=1}^{n^\gamma}\frac{k^{-1-2\alpha (r-\frac{1}{2})}}{1+C_z\left(\frac{k}{n^\gamma}\right)^\alpha},
    &&&
    A_q^n\overset{\mathrm{def}}{=}\sum\limits_{k=1}^{n^\gamma}\frac{k^{-1-2\alpha (r-\frac{1}{2})}}{\left(1+C_z\left(\frac{k}{n^\gamma}\right)^\alpha\right)^2}, 
    &&&
    B^n_q\overset{\mathrm{def}}{=}\sum\limits_{k=1}^{n^\gamma}\frac{k^{-1-2\alpha (r-1)}}{\left(1+C_z\left(\frac{k}{n^\gamma}\right)^\alpha\right)^2}.
\end{align}
We remind that, similarly to the discussion in Appendix \ref{App:mm}, the sequences $A^n_m$ and $A^n_q$ admit identical limits as $n\rightarrow\infty$ : $A^\infty_q=A^\infty_m$. The consequences of this identity are expounded below. We now focus on ascertaining the scalings of the order parameters $m,\, q$. These scalings depend on the regime considered.

\paragraph{Regularized regime $\ell<\alpha$}
In the case $\ell<\alpha$, we have $\gamma=\ell/\alpha <1$. Then the expansion for $q$ reads
\begin{equation}
    q(1-o(1))=\frac{2}{\pi\rho}\left(\rho-\mathbbm{1}_{r\ge\frac{1}{2}}2C_zA^n_q n^{-\ell}\right) -\mathcal{O}\left(n^{-2\ell\mathrm{min}(r,1)}\right)-\frac{4-\pi}{\pi}C_z n^{-\frac{\alpha-\ell}{\alpha}}\int\limits_{0}^\infty\frac{1}{(1+C_zx^\alpha)^2},
\end{equation}
from which it follows that 
\begin{equation}
    q=\frac{2}{\pi\rho} \left(\rho-\mathbbm{1}_{r\ge\frac{1}{2}}2C_zA^n_q n^{-\ell}\right) -\mathcal{O}\left(n^{-\mathrm{min}\left(2\ell\mathrm{min}(r,1),\frac{\alpha-\ell}{\alpha}\right)}\right).
\end{equation}

Therefore the cosine similarity $\eta$ admits the following expression
\begin{equation}
\label{eq:app:ridge:eta}
    \eta=\frac{m^2}{\rho q}=\frac{\frac{2}{\pi\rho} \rho^2\left[1-\mathbbm{1}_{r\ge\frac{1}{2}}\frac{C_zA^n_m}{\rho}n^{-\ell} +\mathcal{O}\left(n^{-2\ell r}\right)\right]^2}{\frac{2}{\pi\rho} \rho^2\left[1-\mathbbm{1}_{r\ge\frac{1}{2}}\frac{2C_zA^n_q}{\rho} n^{-\ell} -\mathcal{O}\left(n^{-\mathrm{min}\left(2\ell\mathrm{min}(r,1),\frac{\alpha-\ell}{\alpha}\right)}\right)\right]}=1-\mathcal{O}\left(n^{-\mathrm{min}\left(2\ell\mathrm{min}(r,1),\frac{\alpha-\ell}{\alpha}\right)}\right),
\end{equation}
where we used that $A^n_q$ and $A^n_m$ share the same limit. Finally, the scaling for the misclassification error can be accessed:
\begin{align}
    \epsilon_g\sim \sqrt{1-\eta}\sim n^{-\frac{1}{2}\mathrm{min}\left(2\ell\mathrm{min}(r,1),\frac{\alpha-\ell}{\alpha}\right)},
\end{align}
which is equation \eqref{eq:ridge_exponent}.
\begin{remark}
Note that only the classification error \eqref{eq:error} tends to zero, while neither the MSE betwen the label $y=\sign( {\theta^{ \star \top}}\psi(x))$ and the pre-activation linear predictor $\hat{w}^\top\psi(x)$ (below denoted $\mathrm{MSE}_1$) nor the MSE between the teacher and student preactivations $\theta^{\star \top}\psi(x)$, $\hat{w}^\top\psi(x)$ (below denoted $\mathrm{MSE}_2$) tend to zero:
\begin{align}
&\mathrm{MSE}_1=\mathbb{E}_{\psi(x)}\left(\sign( {\theta^{ \star \top}}\psi(x))-\hat{w}^\top\psi(x)\right)^2=1+q-2m\sqrt{\frac{2}{\pi\rho}}\xrightarrow[]{n\rightarrow \infty}1-\frac{2}{\pi},\\
    &\mathrm{MSE}_2=\mathbb{E}_{\psi(x)}\left(\theta^{ \star \top}\psi(x)-\hat{w}^\top\psi(x)\right)^2=\rho+q-2m\xrightarrow[]{n\rightarrow \infty} \rho \left(1-\sqrt{\frac{2}{\pi\rho}} \right)^2.
\end{align}
Note also the identity between the order parameter $\hat{r}_2$ and $\mathrm{MSE}_1$: $\hat{r}_2=\mathrm{MSE}_1$, discussed in section \ref{sec:ell2} in the main text. 
\end{remark}

\paragraph{Unregularized regime $\ell>\alpha$} If $\ell>\alpha$, then $\gamma=1$
 and 
 
 \begin{equation}
     q\left(1-C_z\int\limits_{0}^\infty\frac{1}{(1+C_zx^\alpha)^2} \right)=\frac{2}{\pi} -\frac{4-\pi}{\pi}C_z\int\limits_{0}^\infty\frac{1}{(1+C_zx^\alpha)^2} -\mathcal{O}\left(n^{-2\alpha\mathrm{min}(r,\frac{1}{2})}\right).
 \end{equation}
 Hence
 \begin{equation}
     \eta=\frac{m^2}{\rho q}\xrightarrow[]{n\rightarrow\infty} \frac{1-C_z\int\limits_{0}^\infty\frac{1}{(1+C_zx^\alpha)^2}}{1-\frac{4-\pi}{2}C_z\int\limits_{0}^\infty\frac{1}{(1+C_zx^\alpha)^2}}\ne 1.
 \end{equation}
Since $\eta$ is now bounded away from $1$, the misclassification error $\epsilon_g$ fails to go to $0$ asymptotically, and plateaus to a finite value:
 \begin{equation}
      \epsilon_g=\mathcal{O}(1),
 \end{equation}
 which is equation \eqref{eq:plateau}. This plateau is attributable to ridge classifiers overfitting the binary labels, see Appendix \ref{App:double_descent} for further discussion.

\hc{
\subsection{Rates for noisy ridge classification}
In this seubsection, we provide briefly the derivation of the optimal rate for noisy ridge classification, defined by \eqref{eq:noisy_y}, which is plotted in Fig.\,\ref{app:Steinwart:fig:noisy_real} of the main text. Introducing the noise \eqref{eq:noisy_y} in the equations of \cite{Loureiro2021CapturingTL} and mapping to our setting along identical lines as Appendix \ref{App:mm}, the self-consistent equations read
\begin{align}
\label{eq:app:ridge:noisy_g3m_generic}
    \begin{cases} m=\hat{m} p~\mathrm{tr}\frac{\Sigma\theta^\star\theta^{\star \top}\Sigma}{n\lambda+p\hat{V}\Sigma}\\
    q=p\mathrm{tr}\frac{p\hat{m}^2\Sigma\theta^\star\theta^{\star \top}\Sigma^2+\hat{q}\Sigma^2}{(n\lambda+p \hat{V}\Sigma)^2}\\
    V=\mathrm{tr}\frac{\Sigma}{n\lambda+p\hat{V}\Sigma}\\
    \end{cases},
    &&\begin{cases}
    &\hat{V}=\frac{\frac{n}{p}}{1+V}\\
    &\hat{q}=\frac{n}{p}\frac{1+q-2m\sqrt{\frac{2}{\pi(\rho+\sigma^2)}}}{(1+V)^2}=\frac{\hat{V}^2}{\frac{n}{p}}\left(1+q-2m\sqrt{\frac{2}{\pi(\rho+\sigma^2)}}\right)\\
    &\hat{m}=\sqrt{\frac{2}{\pi(\rho+\sigma^2)}}\frac{\frac{n}{p}}{1+V}=\sqrt{\frac{2}{\pi(\rho+\sigma^2)}}\hat{V}.
    \end{cases}
\end{align}
This differs from \eqref{eq:app:ridge:g3m_generic} by a the replacement of $\rho$ by $\rho+\sigma^2$, simply reflecting the fact that the teacher $L^2(\mathcal{X})$ squared norm should now include the contribution of the additive noise.\\

In the \textit{regularized regime} $\ell<\alpha$,the expression \eqref{eq:app:ridge:eta} for the cosine-similarity $\eta$ becomes
\begin{equation}
    \eta=\frac{m^2}{\rho q}=\frac{\frac{2}{\pi(\rho+\sigma^2)} \rho^2\left[1-\mathbbm{1}_{r\ge\frac{1}{2}}\frac{C_zA^n_m}{\rho}n^{-\ell} +\mathcal{O}\left(n^{-2\ell r}\right)\right]^2}{\frac{2}{\pi(\rho+\sigma^2)} \rho^2\left[1-\mathbbm{1}_{r\ge\frac{1}{2}}\frac{2C_zA^n_q}{\rho} n^{-\ell} -\mathcal{O}\left(n^{-\mathrm{min}\left(2\ell\mathrm{min}(r,1),\frac{\alpha-\ell}{\alpha}\right)}\right)\right]}=1-\mathcal{O}\left(n^{-\mathrm{min}\left(2\ell\mathrm{min}(r,1),\frac{\alpha-\ell}{\alpha}\right)}\right).
\end{equation}
Therefore the excess error reads
\begin{align}
    \epsilon_g-\epsilon_g^\infty&=\frac{1}{\pi}\mathrm{cos}^{-1}\left(\sqrt{\frac{\rho}{\rho+\sigma^2}\eta}\right)-\frac{1}{\pi}\mathrm{cos}^{-1}\left(\sqrt{\frac{\rho}{\rho+\sigma^2}}\right)\nonumber\\
    &=\begin{cases}
    \frac{1}{\sqrt{2}\pi}\left(
    \sqrt{\frac{\sigma^2}{\rho+\sigma^2}+(1-\eta)}-\sqrt{\frac{\sigma^2}{\rho+\sigma^2}}\right)& \mathrm{if}~ \sigma^2\ll 1\\
    \frac{1}{2\pi}\frac{\rho+\sigma^2}{\rho}(1-\eta) & \mathrm{if}~ \sigma^2\sim 1
    \end{cases}\\
    &\sim
    1-\eta =\mathcal{O}\left(n^{-\mathrm{min}\left(2\ell\mathrm{min}(r,1),\frac{\alpha-\ell}{\alpha}\right)}\right).
\label{eq:app:ridge:noisy_rate}
\end{align}
Note that the noisy rate \eqref{eq:app:ridge:noisy_rate} corresponds to twice the noiseless rate \eqref{eq:ridge_exponent}.\\

In the \textit{unregularized regime }$\ell>\alpha$, the discussion in the noiseless setting carries through and the excess error also asymptotically saturates to a non-zero limit
\begin{equation}
    \epsilon_g-\epsilon_g^\infty=\mathcal{O}(1)
\end{equation}

\begin{figure}[h]
    \centering
    \includegraphics[scale=0.47]{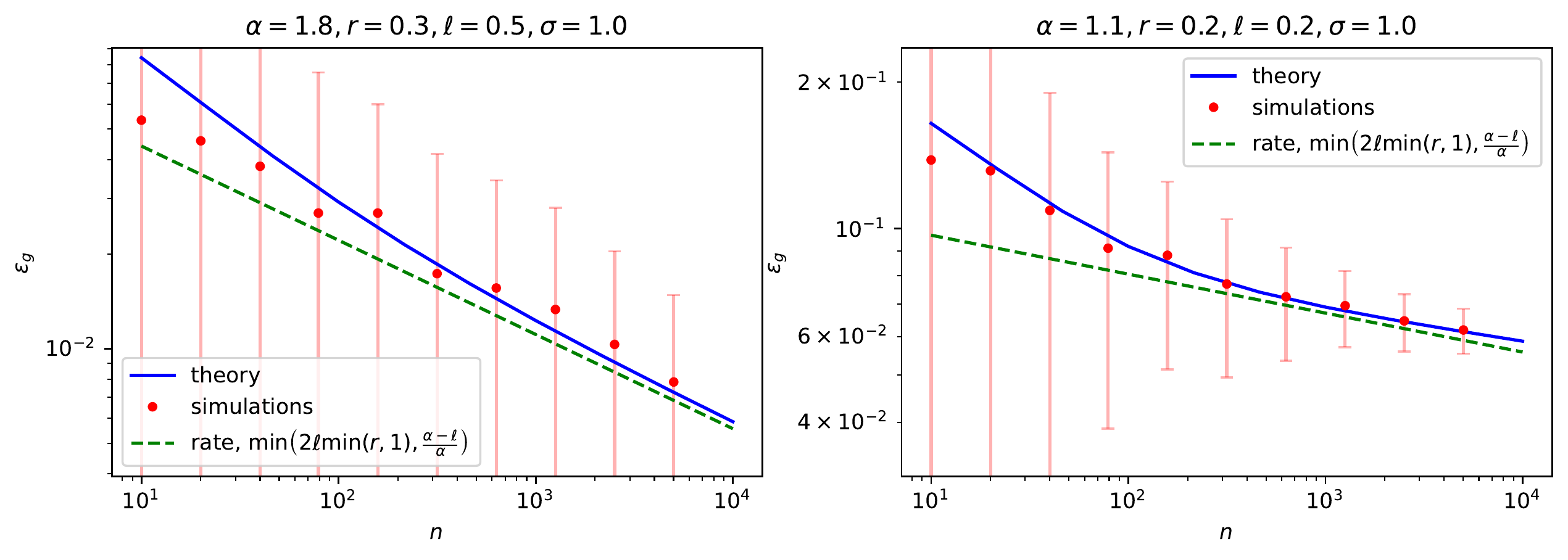}
    \caption{Misclassification error $\epsilon_g$ for ridge classification on synthetic Gaussian data, as specified in \eqref{eq:model_def} and \eqref{eq:noisy_y}, for different source/capacity coefficients $\alpha,r$, for a regularization $\lambda=n^{-\ell}$ and noise strength $\sigma$. In blue, the solution of the closed set of equations \eqref{eq:app:reg:loureiro_g3m_reg} used in the characterization \eqref{eq:error} for the misclassification error, using the \texttt{g3m} package \cite{Loureiro2021CapturingTL}. The dimension $p$ was cut-off at $10^4$. Red dots corresponds to simulations using the \texttt{scikit-learn SVC} package and averaged over $40$ instances, for $p=10^4$. The green dashed line indicates the power-law rate \eqref{eq:app:ridge:noisy_rate}. }
    \label{fig:app:l2noisy}
\end{figure}
}

\section{Remark on the overfitting regime of ridge classification}
\label{App:double_descent}
In this Appendix, we provide further discussion on the saturation of the error \eqref{eq:error} to a finite value in the effectively unregularized regime of ridge classification \eqref{eq:plateau}, which defies the common intuition that for large enough sample complexity $n$, ridge classification should be able to generalize almost perfectly. In this section, we first provide an analytical reminder of why the error does go to zero for finite-dimensional feature spaces ($p<\infty,\,n\gg1$), before discussing what differs in the setup at hand ($1\ll n\ll p=\infty$). We argue that while in the first limit a double-descent phenomenon is observed, the second limit always corresponds to a learning regime located in the plateau/valley following the first descent. In the following, we consider directly the $\lambda=0$ case for the sake of simplicity.

\begin{figure}
    \centering
    \includegraphics[scale=0.48]{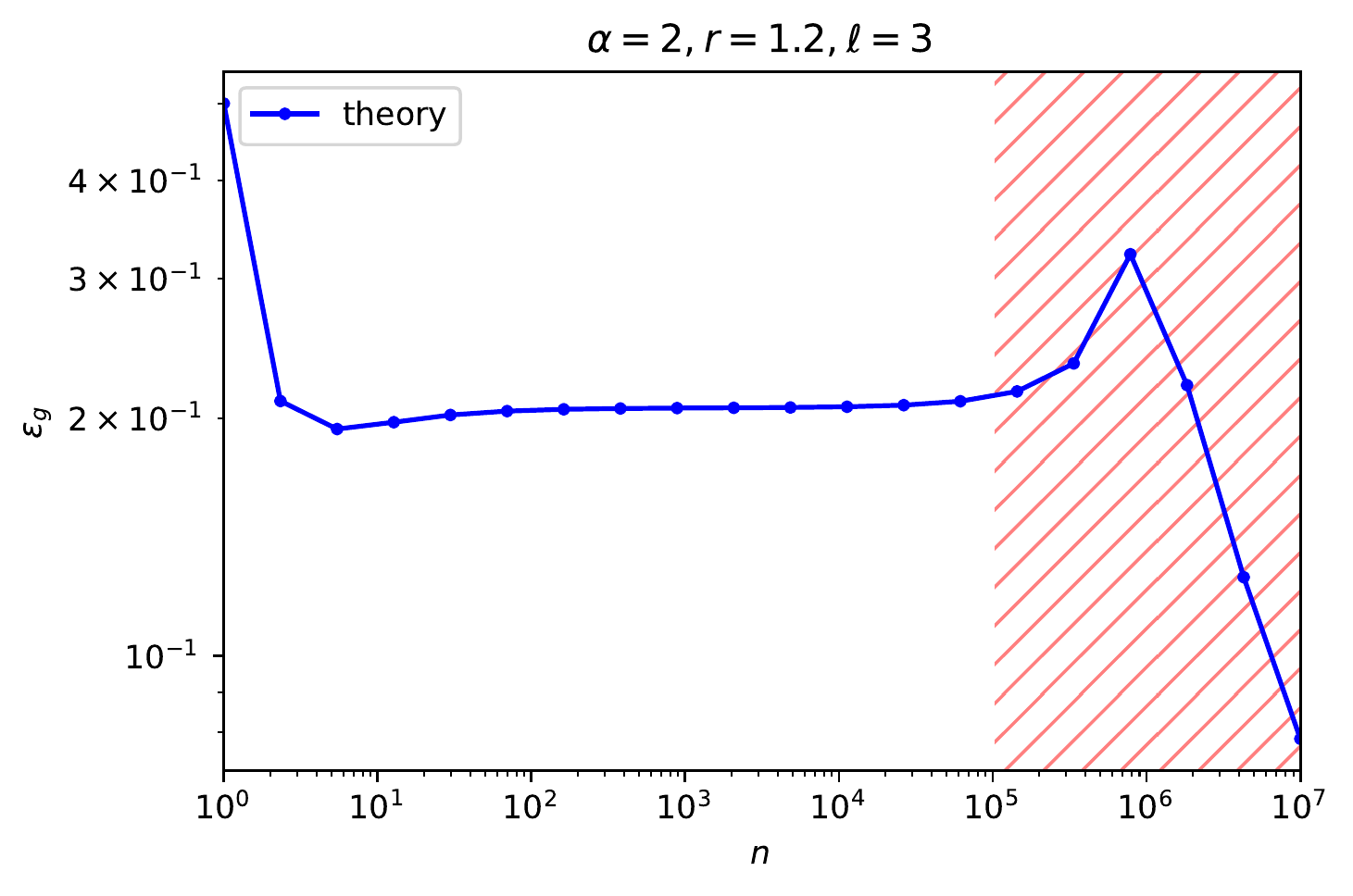}
    \caption{Misclassification error $\epsilon_g$ for ridge classification on artificial Gaussian data, as specified in \eqref{eq:model_def}, for source $r=1.2$, capacity $\alpha=2$, at regularization $\lambda=n^{-3}$. The dimension was cut-off to $p=10^6$. The blue curve corresponds to the numerical solution of the equations \eqref{eq:g3m_ridge} using the \texttt{g3m} package \cite{Loureiro2021CapturingTL}. While for sample complexities comparable or larger than the dimension $n\ge p$ a second descent, followed by a drop of the error to zero, is observable, in the $1\ll n\ll p=\infty$ limit considered in the present work the second descent, which happens for $n\gtrapprox p$ (red hashed region), is no longer observed. Instead, only the plateau following the first descent is observed, which corresponds to the saturation effect of Fig.\,\ref{fig:plateau}}
    \label{fig:app:double_descent}
\end{figure}

\subsection{The finite dimensional setup} 
\label{app:double:subsec:finite_sim}
We first sketch up an brief analysis for the limit of finite feature space dimension $\infty> p\gg 1$ setting, with large sample complexity $n\gg p$. Naming $X\in\mathbb{R}^{n\times p}$ (with the $\mu^{\mathrm{th}}$ row of $X$ being $\psi(x^\mu)$) the matrix of the data in feature space and $y=\mathrm{sign}(X\theta^\star)\in\mathbb{R}^n$ the corresponding vector of stacked labels, the ridge estimator can be compactly written as
\begin{equation}
     \hat{w}=(X^TX)^{-1}X^T\mathrm{sign}(X\theta^\star).
\end{equation}
In the $n\xrightarrow{}\infty $ limit
\begin{align}
    \frac{1}{n}X^TX=\Sigma+\mathcal{O}\left(\frac{1}{\sqrt{n}}\right),
    && \frac{1}{n}X^T\mathrm{sign}(X\theta^\star)=\mathbb{E}_x^{\mathcal{N}(0,\Sigma)}\left[x\mathrm{sign}(x\cdot\theta^\star)\right]+\mathcal{O}\left(\frac{1}{\sqrt{n}}\right).
\end{align}
 Hence 
\begin{align}
    \hat{w}&\approx \Sigma^{-1}\mathbb{E}_x^{\mathcal{N}(0,\Sigma)}\left[x\mathrm{sign}(\theta^{\star T}x)\right]\nonumber\\
    &\approx \Sigma^{-1} \Sigma^{\frac{1}{2}}\mathbb{E}_x^{\mathcal{N}(0,1)}\left[x\mathrm{sign}(x^T\Sigma^{\frac{1}{2}}\theta^{\star})\right]\nonumber\\
    &\approx \Sigma^{-\frac{1}{2}}\mathbb{E}_{x_\parallel}^{\mathcal{N}(0,1)}\left[x_\parallel\mathrm{sign}(x_\parallel)\right]\frac{\Sigma^{\frac{1}{2}}\theta^{\star}}{||\Sigma^{\frac{1}{2}}\theta^{\star}||_2},
\end{align}
where we noted $x_\parallel$ the component of $x$ parallel to $\theta^\star$. So
\begin{equation}
    \hat{w}=\sqrt{\frac{2}{\pi}}\frac{\theta^{\star}}{||\Sigma^{\frac{1}{2}}\theta^{\star}||_2}\propto \theta^\star.
\end{equation}
This means in particular the direction of the teacher is perfectly recovered, and that the prediction error $\epsilon_g$ \eqref{eq:error} goes to zero in the $p\ll n\rightarrow \infty$ limit.\\

\subsection{Infinite dimensional feature space}

We now argue why this discussion ceases to hold in the setup of interest $n\ll p=\infty$ (plateau reached after the first descent, see Fig.~\ref{fig:app:double_descent}). In this limit, loosely written,
\begin{align}
    \hat{w}&\approx \left(\Sigma^{-1}+\mathcal{O}\left(\frac{1}{\sqrt{n}}\right)\right)\left(\mathbb{E}_x^{\mathcal{N}(0,\Sigma)}\left[x\mathrm{sign}(\theta^{\star T}x)\right]+\mathcal{O}\left(\frac{1}{\sqrt{n}}\right)\right)\nonumber\\
    &=\sqrt{\frac{2}{\pi}}\frac{\theta^{\star}}{||\Sigma^{\frac{1}{2}}\theta^{\star}||_2}+\mathcal{O}\left(\sqrt{\frac{p}{n}}\right) \ne \mathrm{cte.}\times\theta^\star.
\end{align}
The last term comes from the sum of $p$ random terms of order $1/\sqrt{n}$ entailed by the matrix multiplication. In particular, this implies that the teacher fails to be perfectly recovered by ERM in this limit, causing the misclassification error to plateau to a finite limit, see equation \eqref{eq:plateau}. This is due to the fact that in an infinite-dimensional space, the ridge ERM \eqref{eq:ridge_risk}  always has enough dimensions to overfit the dataset. Another way to look at this limit is in the framework of the well-known \textit{double-descent} phenomenon which occurs in the finite dimensional limit $p<\infty$ discussed in subsection \ref{app:double:subsec:finite_sim}. In the $p=\infty$ limit, the second descent, which commences at $n\approx p$, happens at infinite sample complexity and is thus not observed for finitely large $n$. In fact, the $1\ll n\ll p=\infty$ limit always correspond to the plateau following the first descent, see Fig.\,\ref{fig:app:double_descent}.

\section{Crossovers in noisy kernel classification }
\label{App:Steinwart}
\hc{

In this appendix, we briefly discuss the \textit{noisy} setup where the target function $f^\star$ is corrupted by a Gaussian noise of variance $ \sigma^2$, i.e.
\begin{equation}
  y^\mu=\mathrm{sign}\left(f^\star(x^\mu)+\sigma\mathcal{N}(0,1)\right) .
  \label{eq:noisy_y}
\end{equation}
In this setting, the misclassification error \eqref{eq:error} no longer tends asymptotically to $0$, but instead converges to a positive value dictated by the noise strength
\begin{equation}
    \epsilon_g\xrightarrow[]{n\rightarrow\infty }\epsilon_g^\infty=\frac{1}{\pi}\mathrm{arccos}\left(\sqrt{\frac{\rho}{\rho+\sigma^2}}\right).
\end{equation}
Observe that the residual error $\epsilon_g^\infty$ corresponds to the smallest achievable $0/1$ risk, and is achieved by the teacher $f^\star$. In contrast to the noiseless setting \eqref{eq:gt_function}, the stochasticity in the generation of the labels \eqref{eq:noisy_y} induces a non-zero error even for $f^\star$.  Numerical experiments presented in Fig.\,\ref{app:Steinwart:fig:crossover} show that the rate of decay of the excess misclassification error $\epsilon_g-\epsilon_g^\infty$, when optimized over the regularization strength $\lambda$, transitions from the noiseless rate \eqref{eq:scaling_mm} to a $\alpha/(1+\alpha)$ rate as the sample complexity $n$ is increased. Note that interestingly the rate $\alpha/(1+\alpha)$ has been reported in \cite{Steinwart2008SupportVM} as an upper bound for the excess misclassification error, although it is possible to show that the noisy setup \eqref{eq:noisy_y} does \textit{not} satisfy the conditions of their corresponding theorem. This crossover phenomenon is highly similar to the one observed in \cite{Cui2021GeneralizationER} for kernel regression.

\begin{figure}
    \centering
    \includegraphics[scale=0.48]{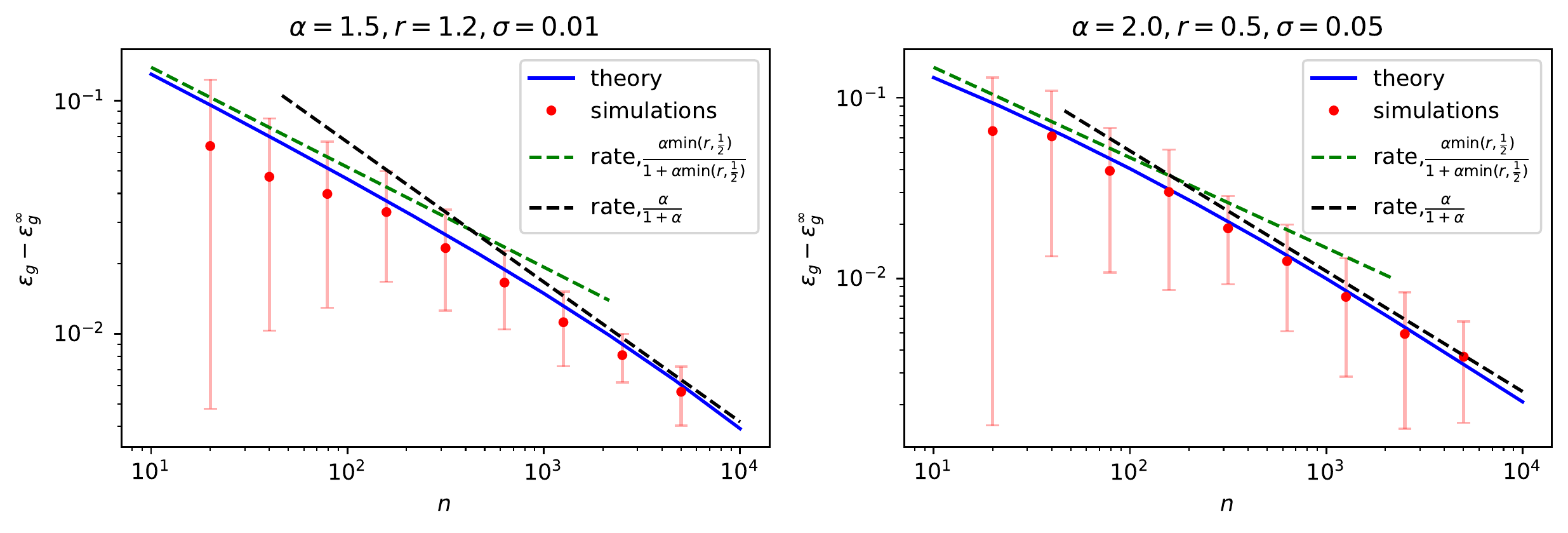}
    \caption{Excess misclassification error $\epsilon_g-\epsilon_g^\infty$ for max-margin classification on synthetic Gaussian data corrupted by a Gaussian noise of variance $\sigma^2$, as specified in \eqref{eq:model_def}, for different source/capacity coefficients $\alpha,r$, for optimal $\lambda^\star$. In blue, the solution of the closed set of equations \eqref{eq:g3m_mm} used in the characterization \eqref{eq:error} for the misclassification error, using the \texttt{g3m} package \cite{Loureiro2021CapturingTL}. The dimension $p$ was cut-off at $10^4$. Red dots corresponds to simulations using the \texttt{scikit-learn SVC} package averaged over $40$ instances, for $p=10^4$. Optimization over $\lambda$ was performed using cross validation, with the help of the python \texttt{scikit-learn GridSearchCV} package. The green dashed line indicates the power-law rate \eqref{eq:scaling_mm} derived in this work; the black dashed line indicates the classical optimal rate $\alpha/(1+\alpha)$ \cite{Steinwart2008SupportVM}. }
    \label{app:Steinwart:fig:crossover}
\end{figure}
}

\begin{figure}[ht!]
    \centering
    \includegraphics[scale=0.48]{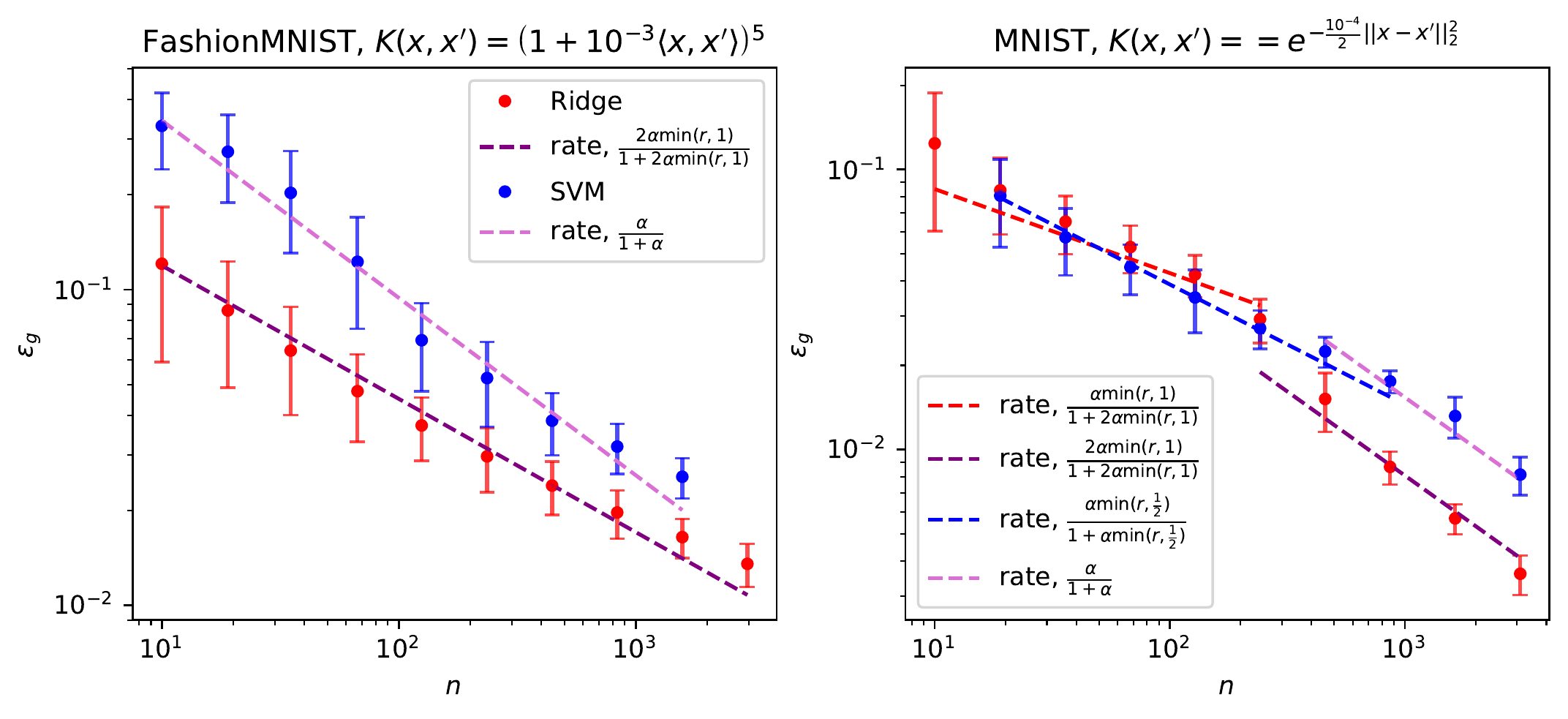}
    \caption{Dots: Misclassification error $\epsilon_g$ of kernel classification on Fashion MNIST with a polynomial kernel (left) and MNIST with an RBF kernel, for optimally regularized SVM (blue) and optimally regularized ridge classification (red), using respectively the python \texttt{scikit-learn SVC} and \texttt{KernelRidge} packages. Dashed lines: Theoretical decay rates for the error $\epsilon_g$ (blue: \eqref{eq:scaling_mm}, light purple :\cite{Steinwart2007FastRF}, red: \eqref{eq:opt_rates_ridge}, purple: \eqref{eq:app:ridge:noisy_rate}), computed from empirically estimated capacity $\alpha$ and source $r$ coefficients (see section \eqref{sec:real} and Appendix\,\ref{App:real} for details). These coefficients were estimated to $\alpha\approx 1.28,\,r\approx 0.28$  for FashionMNIST using the polynomial kernel and $\alpha\approx 1.5,\,r\approx 0.5$ for MNIST using the RBF kernel, see Appendix\,\ref{App:real}. }
    \label{app:Steinwart:fig:noisy_real}
\end{figure}

 \hc{Finally, we note that for some real datasets the learning curves are satisfactorily described by the noisy rate $\alpha/(1+\alpha)$\cite{Steinwart2008SupportVM} (see also the discussion in section \ref{sec:maxmargin}). This is for instance the case for Fashion MNIST when a polynomial kernel SVM is employed, see Fig.\,\ref{app:Steinwart:fig:noisy_real}. Interestingly, the rate which describes the learning curve of an optimally regularized ridge classifier in this setting is also the \textit{noisy} rate \eqref{eq:app:ridge:noisy_rate}(see Appendix\,\ref{App:ridge}), rather than the noiseless rate \eqref{eq:opt_rates_ridge}. For MNIST classified using an RBF kernel, both the noiseless rates \eqref{eq:scaling_mm},\eqref{eq:opt_rates_ridge} and the noisy rates $\alpha/(1+\alpha)$,\eqref{eq:app:ridge:noisy_rate} are observed, with a crossover phenomenon during which the learning curve transitions from the former to the latter.}

\section{Details on real data-sets}
\label{App:real}
\subsection{Measuring the capacity and source of real datasets}

\begin{figure}
    \centering
    \includegraphics[scale=0.47]{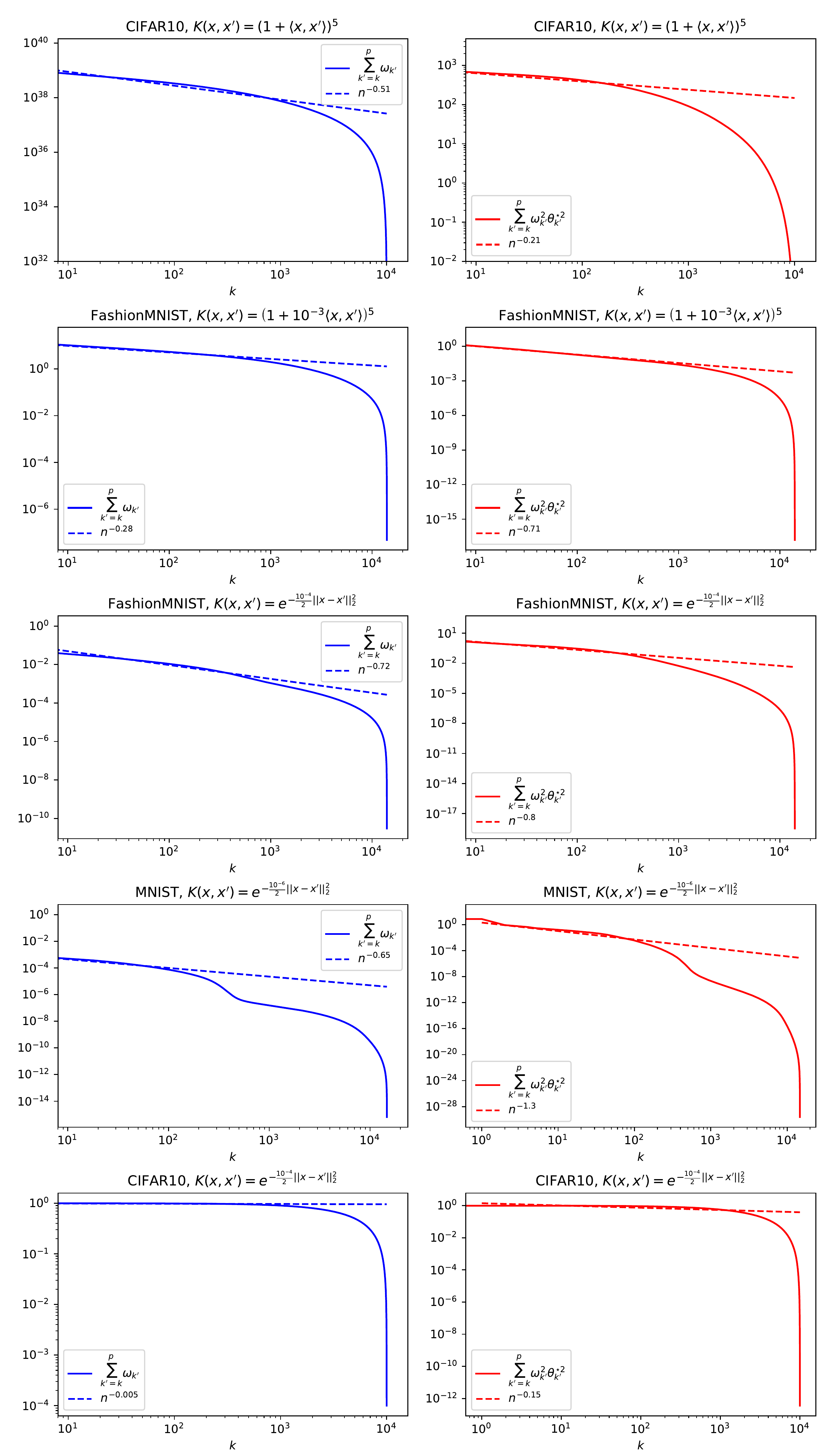}
    \caption{Cumulative functions \eqref{eq:app:cumu} for $10^4$ CIFAR 10 sampled at random, for a RBF kernel (top) and a polynomial kernel (bottom). The slopes fitted using least-squares linear regression on the regions the curves are qualitatively resembling power-laws are represented in dashed lines.}
    \label{fig:app:measure_coeffs}
\end{figure}

In this section we provide details on the experiments on real data (section \ref{sec:real}, Fig.\,\ref{fig:cifar}), which can also be found in \cite{spigler2019asymptotic,bordelon2020,Cui2021GeneralizationER}. Consider a real dataset $\mathcal{D}_m=\{x^\mu,y^\mu\}^m_{\mu=1}$ of size $m$. The data distribution $\nu$ defined in section \ref{sec:setting} is then the empirical distribution over $\mathcal{D}_m$:
\begin{equation}
    \nu(\cdot)=\frac{1}{m}\sum\limits_{\mu=1}^m \delta(\cdot-x^\mu).
\end{equation}
Then the definitions of the feature map $\psi$ \eqref{eq:def_psi} and of the feature covariance $\Sigma$ \eqref{eq:def_Sigma} admit the simple rewriting
\begin{align}
\label{eq:app:real:matrix_notations}
    \frac{1}{m}G\Psi=\Psi \Sigma, && \Sigma=\frac{1}{m}\Psi^\top\Psi,
\end{align}
provided we introduce the Gram matrix $G_{\mu, \nu}=K(x^\mu,x^\nu)\in\mathbb{R}^{m\times m}$ and the matrix of horizontally stacked features $\Psi_{\mu k}=\psi(x^\mu)_k \in\mathbb{R}^{m\times m}$. Note that for real data, one has that the dimension of the feature space $p$ is equal to the total size of the dataset $m$. A teacher $\theta^\star$ providing perfect classification $ y=\sign(\Psi\theta^\star)$ can be found e.g. by performing max margin classification on $(\Psi,y)$. While the method to fit a perfect classifier $\theta^\star$ is not unique, we observed that using logistic classifiers or regularized hinge classifiers instead of max-margin classifiers did not significantly impact the measured source coefficient. \\

From the eigenvalues $\{\eig_k\}_{k=1}^p$ of the covariance $\Sigma$ \eqref{eq:app:real:matrix_notations} and the components of the teacher $\{\theta^\star_k\}_{k=1}^p$, the capacity and source $\alpha,\,r$ \eqref{eq:model_def} can be estimated. From \eqref{eq:model_def}, the following scalings should hold:
\begin{align}
\label{eq:app:cumu}
    \sum\limits_{k'=k}^p \eig_{k'}\sim k^{1-\alpha},
    &&\sum\limits_{k'=k}^p \eig_{k'}\theta_{k'}^{\star 2}\sim k^{-2\alpha r}.
\end{align}
These curves are presented in Fig.\,\ref{fig:app:measure_coeffs} for the dataset comprised of $10^4$ randomly samples CIFAR 10 images, see section \ref{sec:real} of the main text.
Note that we estimate the cumulative functions \eqref{eq:app:cumu} following \cite{spigler2019asymptotic,Cui2021GeneralizationER} because the summation allows to smoothen out the curves, thus permitting a relatively more precise evaluation. Because the feature space is finite, the functions \eqref{eq:app:cumu} fail to be exact power-laws, and exhibit in particular a sharp drop for $n$ approaching $m$. Nonetheless, we identify for each curve a region of indices $k$ for which the curve looks qualitatively like a power-law, and fit the curves by a power-law using least-squares linear regression to finally extract the coefficients $\alpha, r$ therefrom. These fits are shown in Fig.\,\ref{fig:app:measure_coeffs} for the reduced CIFAR 10 dataset discussed in the main text (see section \ref{sec:real}), for an RBF kernel with inverse variance $10^{-7}$ and a polynomial kernel of degree $5$. The capacity and source were estimated to be $\alpha\approx 1.16,\,r\approx 0.10$ for the RBF kernel and $\alpha\approx 1.51,\,r\approx 0.07$ for the polynomial kernel.

\subsection{Details on numerical simulations}
In this subsection, we provide further details on the simulations presented in Fig.\,\ref{fig:cifar} of the main text. For each sample complexity $n$, a subset of size $n$ was randomly sampled without replacements from $\mathcal{D}_m$. Like in \cite{Canatar2021SpectralBA, Loureiro2021CapturingTL, Cui2021GeneralizationER}, the whole dataset $\mathcal{D}_m$ was used as a test set. The max-margin simulations in Fig.\,\ref{fig:cifar} were performed using the \texttt{scikit-learn SVC} package at vanishing regularization $\lambda=10^{-5}$, and averaged over $50$ realizations of the training set. The ridge simulations in Fig.\,\ref{fig:cifar} were realized using the \texttt{scikit-learn KernelRidge} package, with the optimal $\lambda$ estimated using \texttt{scikit-learn GridSearchCV}'s default $5$-fold cross validation routine over a grid $\lambda\in \{0\}\cup (10^{-10},10^5)$, with logarithmic step size $0.026$. The misclassification error was also averaged over $50$ realizations of the training set.

\section{Comparison to the classical rates}
\label{App:loose}

In this Appendix, we provide a detailed comparison of the presently reported rate for noiselesss max-margin classification \eqref{eq:scaling_mm} and the rate reported in \cite{Steinwart2008SupportVM} (equation $7.53$, following from Theorem $7.23$ and Lemma $A.1.7$), in the case of a teacher in the Hilbert space $f^\star\in\mathcal{H}$, for which the latter result holds. We alternatively point the reader to section $5$ of \cite{Vecchia2021RegularizedEO}, where the same bound, and the underlying assumptions thereof, are reminded in concise fashion. We first provide a brief reminder of this upper bound, before proceeding to evaluate it in the present setting \eqref{eq:model_def}, and show that it yields a slower rate than \eqref{eq:scaling_mm}, i.e. that the bound in \cite{Steinwart2008SupportVM} is loose. For completeness, we finally provide a comparison to the rates reported in \cite{Audibert2007FastLR}.

\subsection{Theorem 7.23 in [23]}
In the following, $\ell(y,a)=\mathrm{max}(0,1-ya)$ designates the hinge loss, and the script cl refers to the clipped value \cite{Steinwart2008SupportVM}.
Suppose that
\begin{enumerate}[label=A.\arabic*]
\item $\underset{w}{\mathrm{inf}}\mathbb{E}_{\psi(x),y}\ell(y,w^\top \psi(x))=\mathbb{E}_{\psi(x),y}\ell\left(y,f^*(\psi(x))\right)$ where the Bayes estimator $f^*$ is defined as $f^*(\psi(x))=\underset{a}{\mathrm{argmin}}\mathbb{E}_{\psi(x),y}\ell(y,a)$. The expectation refers to the joint distribution of the data $\psi(x)$
 and the labels $y$.
 \item (\textit{Bernstein condition}) There exist $B>0,\,\theta\in[0,1],\,V\ge B^{2-\theta}$ so that for all $w$, $\ell(y,(w^\top \psi(x))^{\mathrm{cl}})\le B~~ \mathrm{a.s.}$ and
\begin{align}
    \mathbb{E}_{\psi(x),y}\left(\left[
    \ell(y,(w^\top \psi(x))^{\mathrm{cl}})-\ell(y,f^*(\psi(x)))
    \right]^2\right)\le 
    & V \Bigg[\mathbb{E}_{\psi(x),y}\Bigg(\ell(y,(w^\top \psi(x))^{\mathrm{cl}})\nonumber \\
    &-\ell(y,f^*(\psi(x)))
    \Bigg)\Bigg]^\theta
\end{align}
\item the spectrum of the covariance $ \Sigma=\mathbb{E}_{\psi(x)}\psi(x)\psi(x)^\top$ has a polynomial decay with rate $\alpha$
\item The approximation error $\mathcal{A}_2( \lambda)=\underset{w}{\mathrm{min}}\mathbb{E}_{\psi(x),y}\ell(y,w^\top \psi(x))+\lambda||w||^2-\underset{w}{\mathrm{inf}}\mathbb{E}_{\psi(x),y}\ell(y,w^\top \psi(x))$ is upperbounded by $\lambda^b$ for $b\in(0,1]$
 \end{enumerate}
 Then with high probability \cite{Steinwart2008SupportVM}
 \begin{equation}
     \epsilon_g-\epsilon_g^*\lesssim n^{-\mathrm{min}\left(\frac{2b}{1+b},\frac{
     \alpha b}{b(2\alpha-1-\alpha\theta+\theta)+1}\right)}.
     \label{eq:app:loose:theorem}
 \end{equation}
 In \eqref{eq:app:loose:theorem}, $\epsilon^*$ refers to the infimum of the classification error over all measurable functions (as opposed to $\epsilon_g$ resulting from ERM \eqref{eq:MM_risk} over linear estimators only).
 
 \subsection{Specialization to the noiseless source/capacity setting}
\begin{figure}
    \centering
    \includegraphics[scale=0.45]{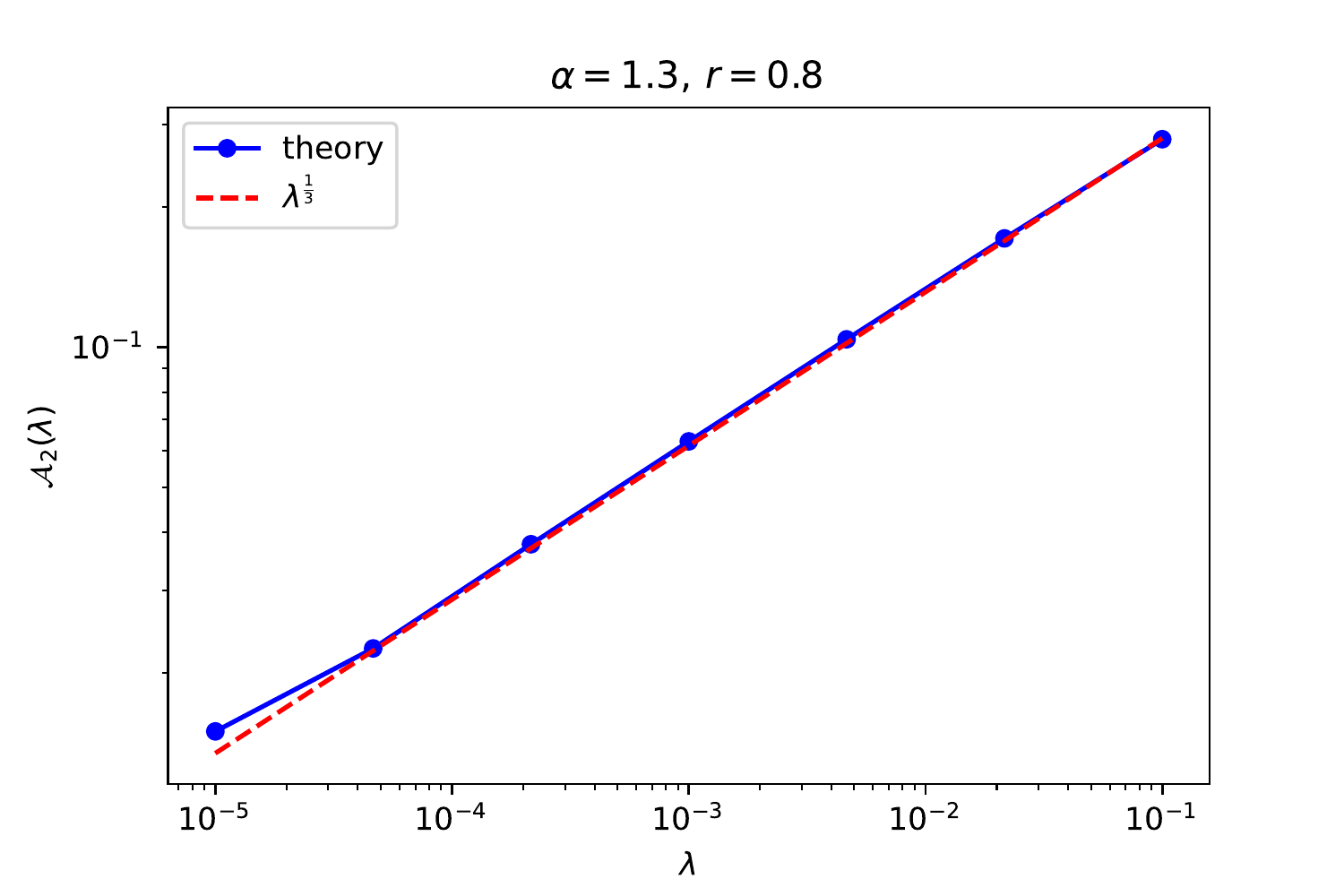}
    \includegraphics[scale=0.45]{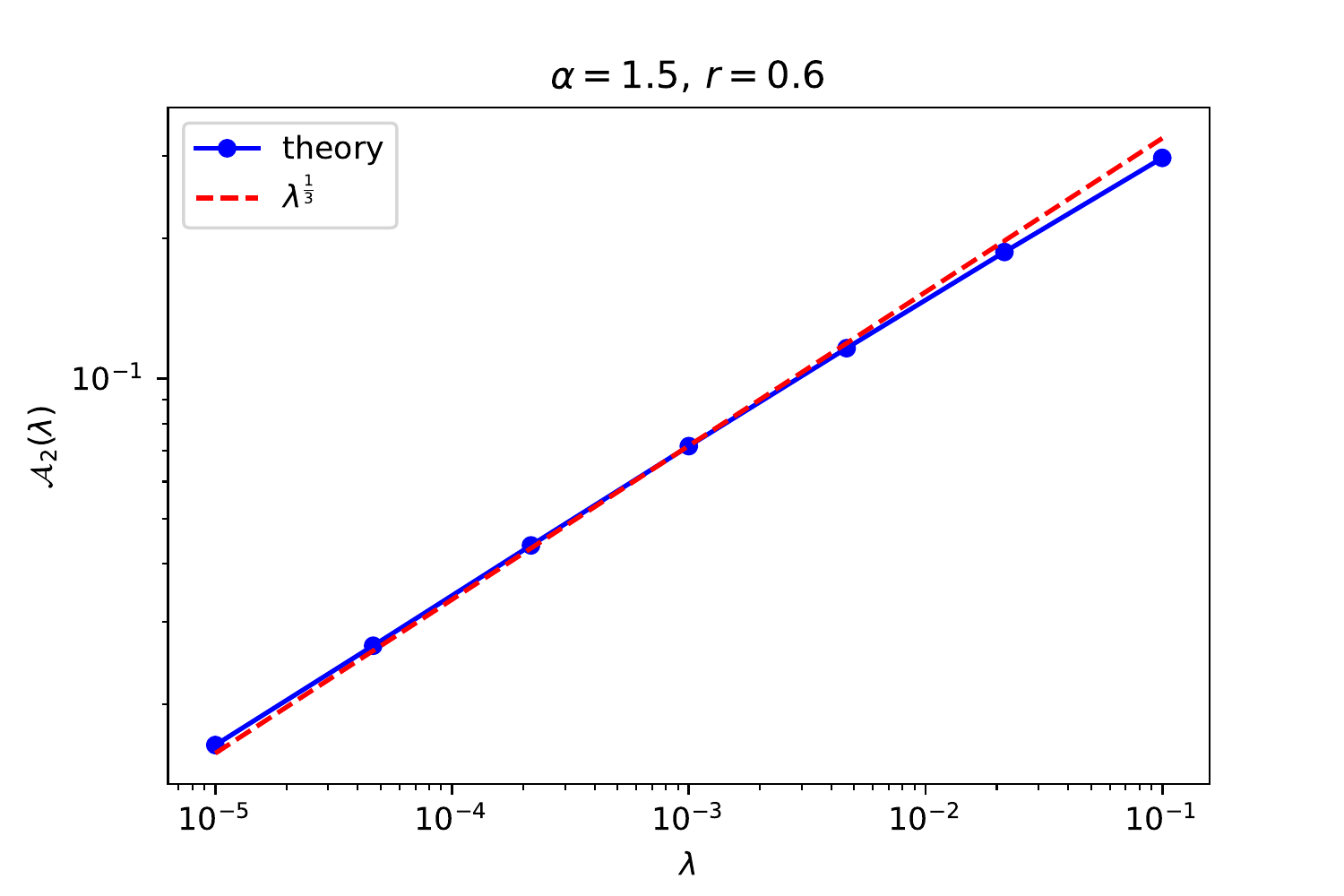}
    \caption{Approximation error $\mathcal{A}_2(\lambda)$ as a function of the regularization strength $\lambda$. In blue, the numerical solution of the self-consistent equations \eqref{eq:g3m_mm} using the \texttt{g3m} package \cite{Loureiro2021CapturingTL}. Red dotted lines represent the power law $\lambda^{\frac{1}{3}}$, which provides a very satisfying fit for all values of capacity/source coefficients $\alpha,r$.}
    \label{fig:A2lambda}
\end{figure}

 We proceed to translate Theorem \eqref{eq:app:loose:theorem} \cite{Steinwart2008SupportVM}  for our setting \eqref{eq:model_def} \eqref{eq:gt_function}, in the case where the teacher $f^\star$ is in $\mathcal{H}$ ($r\ge \frac{1}{2})$, for which assumption (A.1) is readily satisfied. To this end, one must evaluate the parameters $\theta,\,b$ entering in \eqref{eq:app:loose:theorem}. First note that the Bayes estimator $f^*$ assumes the simple form $f^*(x)=\mathrm{sign}(\theta^{\star\top}x)$ in the presence of the teacher $\theta^\star$. As a consequence, in the noiseless setting,
 \begin{equation}
     \mathbb{E}_{\psi(x),y}\ell\left(y,f^*(\psi(x))\right)=\mathbb{E}_{\psi(x)}\ell\left(\mathrm{sign}(\theta^{\star\top}x),\mathrm{sign}(\theta^{\star\top}x))\right)=0.
 \end{equation}
 The Bernstein condition is straightforwardly satisfied with $B=V=2$ and $\theta=1$, since $\ell(1,\cdot^{\mathrm{cl}})$ (resp. $\ell(1,\cdot^{\mathrm{cl}})$) is a piecewise continuous function, constant equal to $0$ on $(1,\infty)$ (resp$(-\infty,-1)$) and to $2$ on $(-\infty,-1)$ (resp. $(1,\infty)$), and linear in between. In our model, the argmin in the definition of $\mathcal{A}_2(\lambda)$ can be approximated first as $w\approx c\times \theta^\star/||\theta^\star||$, i.e. for a $w$ proportional to the teacher. Note that this is an approximation, and in general there exist a small (shrinking with $\lambda$ going to $0$) angle between the true minimizer and the teacher. Typically, $c$ will tend to infinity as $\lambda$ goes to zero. Let us evaluate the proportionality constant $c$:
\begin{align}
    \mathbb{E}_{\psi(x),y}\ell(y,w^\top \psi(x))&\propto \int dx e^{-\frac{1}{2}x^\top \Sigma^{-1}x}\mathrm{max}(0,1-\mathrm{sign}(\theta^{\star\top}x) c\theta^{\star\top} x/||\theta^\star||)\nonumber\\
    &\propto \int\limits_{c\theta^{\star\top}x/||\theta^\star||<1} dx e^{-\frac{1}{2}x^\top \Sigma^{-1}x}(1-\mathrm{sign}(\theta^{\star\top}x) c\theta^{\star\top} x/||\theta^\star||)\nonumber\\
    &\overset{x_\parallel=\theta^{\star\top}x/||\theta^\star||}{\propto}\int  \limits_{-\frac{1}{c}}^{\frac{1}{c}}dx_\parallel e^{-\frac{x_\parallel^2}{2\theta^{\star\top}\Sigma\theta^\star/||\theta^\star||^2}} (1-c|x_\parallel|)\nonumber\\
    &\propto 2\int\limits_{0}^{1} dx_\parallel e^{-\frac{x_\parallel^2}{2c\theta^{\star\top}\Sigma\theta^\star/||\theta^\star||^2}} (1-x)\sim\frac{1}{c}
\end{align}
where in the last line we supposed $c$ large. Then 
\begin{equation}
    \mathcal{A}(\lambda)\approx\underset{c}{\mathrm{min}}\left\{ \frac{A}{c}+\lambda c^2
    \right\}
\end{equation}
for some constant $A$, for $\lambda$ small ($c$ large). Minimizing the function leads to $c\sim \lambda^{-\frac{1}{3}}$, then to $\mathcal{A}(\lambda)\sim \lambda^{\frac{1}{3}}$, i.e. $b=\frac{1}{3}$. We also checked this scaling by numerically solving the self-consistent equations \eqref{eq:g3m_mm} in the large sample complexity limit ($n\gg p$), noticing that the train loss admits the following expression in terms of the order parameters $m,q,\rho,V,\hat{m},\hat{q},\hat{V}$ \cite{Loureiro2021CapturingTL}:
\begin{align}
    \mathcal{A}_2(\lambda)\approx \underset{n\gg p}{\mathrm{lim}}  ~~&2\int\limits_0^\infty \frac{e^{- \frac{u^2}{2}}}{\sqrt{2\pi}}\int\limits_{-\infty}^\infty \frac{e^{-\frac{h^2}{2}}}{\sqrt{2\pi}}\mathrm{max}\left(
    0,
    1-\mathrm{prox}_{V\mathrm{max}(0,1-\cdot)}\left(
    \frac{m}{\sqrt{\rho}}s+\sqrt{q(1-\eta)}h
    \right)
    \right)\nonumber\\
    &+\lambda p\mathrm{tr}\frac{p\hat{m}^2\Sigma\theta^\star\theta^{\star \top}\Sigma+\hat{q}\Sigma}{(\frac{n\lambda}{2}+p \hat{V}\Sigma)^2},
\end{align}
where the proximal map is given by
\begin{align}
\mathrm{prox}_{V\mathrm{max}(0,1-y\cdot)}(\omega)=
    \begin{cases}
    \omega+yV~~\mathrm{if}~~y\omega\le 1-V,\\
    y~~\mathrm{if}~~1-V \le y\omega\le 1,\\
    \omega~~ \mathrm{if}~~y\omega\ge 1.
    \end{cases}.
\end{align}
The function $\mathcal{A}_2(\lambda)$ is plotted for several values of source/capacity coefficients $\alpha,r$ in Fig.\,\ref{fig:A2lambda} and is well captured by a power law with rate $b=\frac{1}{3}$.

Finally, the bound provided by the Theorem \eqref{eq:app:loose:theorem} \cite{Steinwart2008SupportVM} reads (notice that $\epsilon^*_g=0)$
 \begin{equation}
     \epsilon_g-\epsilon_g^*\lesssim n^{-\mathrm{min}\left(\frac{1}{2},\frac{\frac{\alpha}{3}}{\frac{\alpha}{3}+1}\right)}
 \end{equation}
 Note that the rate \eqref{eq:scaling_mm} is always faster:
 \begin{equation}
     n^{-\frac{\frac{\alpha}{2}}{1+\frac{\alpha}{2}}}\ll n^{-\mathrm{min}\left(\frac{1}{2},\frac{\frac{\alpha}{3}}{\frac{\alpha}{3}+1}\right)}.
 \end{equation}
 In other words, the upper bound of \cite{Steinwart2008SupportVM} is loose in the present setting when $f^\star\in\mathcal{H}$. While numerical investigations suggest this is also true for $f^\star\in L^2(\mathcal{X})\setminus\mathcal{H}$, we leave a more detailed comparison to \cite{Steinwart2008SupportVM} in this case to future work.

 

 \subsection{Theorems 3.3 and 3.5 in [25] }
 Finally, we discuss the rates reported in \cite{Audibert2007FastLR} under margin assumptions as \cite{Boucheron2005TheoryOC,Steinwart2007FastRF}. Note that the regression function $\eta\left(\psi(x)\right)\equiv \mathbb{P}(y=1|\psi(x))$ has in the noiseless setting \eqref{eq:gt_function} the compact form $\eta\left(\psi(x)\right)=\sign(\theta^{\star\top}\psi(x))$ and is therefore \textit{not} H\"older class for any $\beta>0$, rendering Theorems $3.3$ and $3.5$ in \cite{Audibert2007FastLR} effectively inapplicable. Note that in addition, the rates would in any case be ambiguous, as both the margin exponent $\alpha$ and the dimension $d$ are infinite in our setting. Finally remark that \cite{Steinwart2007FastRF, Paccolat2021HowIK} provide rates under similar margin conditions in \textit{direct space}, for the particular case of Gaussian kernels. Relating those conditions to the characterization \eqref{eq:model_def} in \textit{feature space} used in the present work is out of the scope of the present manuscript.

\end{document}